%% file: acl2023_dialog.tex
\pdfoutput=1

\documentclass[11pt]{article}

\usepackage{ACL2023}

\usepackage{times}
\usepackage{latexsym}

\usepackage[T1]{fontenc}

\usepackage[utf8]{inputenc}

\usepackage{microtype}

\usepackage{inconsolata}

\usepackage{epsfig}
\usepackage{graphicx}
\usepackage{xcolor}
\usepackage{booktabs}
\usepackage{caption}
\usepackage{makecell}
\usepackage{lipsum}
\usepackage{tabularx}
\usepackage{diagbox}
\usepackage{subfig}
\usepackage[export]{adjustbox}
\usepackage{pifont}

\input{math_commands.tex}

\include{mymacros}

\newcommand{\datasetName}{\textsc{MPChat}\xspace}
\newcommand{\entailmentLabelEntailed}{\textsc{Entailed}\xspace}
\newcommand{\entailmentLabelNotEntailed}{\textsc{Not Entailed}\xspace}
\title{\datasetName: Towards Multimodal Persona-Grounded Conversation}

\author{Jaewoo Ahn\textsuperscript{1} \qquad
        Yeda Song\textsuperscript{1} \qquad
        Sangdoo Yun\textsuperscript{2,1} \qquad
        Gunhee Kim\textsuperscript{1} \qquad
        \\
        \textsuperscript{1}{Seoul National University} \qquad
        \textsuperscript{2}{NAVER AI Lab}
        \\
        {\tt\small \{jaewoo.ahn,yeda.song\}@vision.snu.ac.kr, sangdoo.yun@navercorp.com, gunhee@snu.ac.kr} \\ 
        \small\url{http://vision.snu.ac.kr/projects/mpchat}
}

\begin{document}
\maketitle
\begin{abstract}
    In order to build self-consistent personalized dialogue agents,
    previous research has mostly focused on \textit{textual persona} that delivers personal facts or personalities.
    However, to fully describe the multi-faceted nature of persona,
    image modality can help better reveal the speaker’s personal characteristics and experiences in episodic memory~\citep{Rubin:2003:MC,Conway:2009:Neuropsychologia}.
    In this work, we extend persona-based dialogue to the multimodal domain and make two main contributions.
    First, we present the first multimodal persona-based dialogue dataset named \datasetName,
    which extends persona with both text and images to contain episodic memories.
    Second, we empirically show that incorporating multimodal persona, as measured by three proposed multimodal persona-grounded dialogue tasks (\ie next response prediction, grounding persona prediction, and speaker identification),
    leads to statistically significant performance improvements across all tasks.
    Thus, our work highlights that multimodal persona is crucial for improving multimodal dialogue comprehension,
    and our \datasetName serves as a high-quality resource for this research.
\end{abstract}

\section{Introduction}
\label{sec:introduction}

With the rapid advance of conversational AI systems in recent years, developing self-consistent dialogue agents has been studied much~\citep{Li:2016:ACL,Zhang:2018:ACL}.
Considerable research aims to endow dialogue agents with \textit{persona}, which represents an individual's personality \citep{Zhong:2022:NAACL,Cao:2022:ACL}.
In particular, researchers have exploited \textit{textual description} of persona, for example, in the form of unstructured sentences~\citep{Mazare:2018:EMNLP},
structured key-value attributes (\eg age, gender, location)~\citep{Song:2020:EMNLP} and personality types (\eg Big-Five)~\citep{Mairesse:2007:ACL}.
Therefore, dialogue agents with persona have been found to (1) exhibit greater self-consistency~\citep{Welleck:2019:ACL,Kim:2020:EMNLP,Majumder:2020:EMNLP},
(2) demonstrate awareness of long-term memory~\citep{Xu:2022:ACL,Xu:2022:ACLF,Bae:2022:EMNLPF}, and (3) generate engaging responses instead of non-specific ones~\citep{Zhang:2018:ACL,Mazare:2018:EMNLP}.

\begin{figure}[t]
\begin{center}
    \includegraphics[width=\columnwidth]{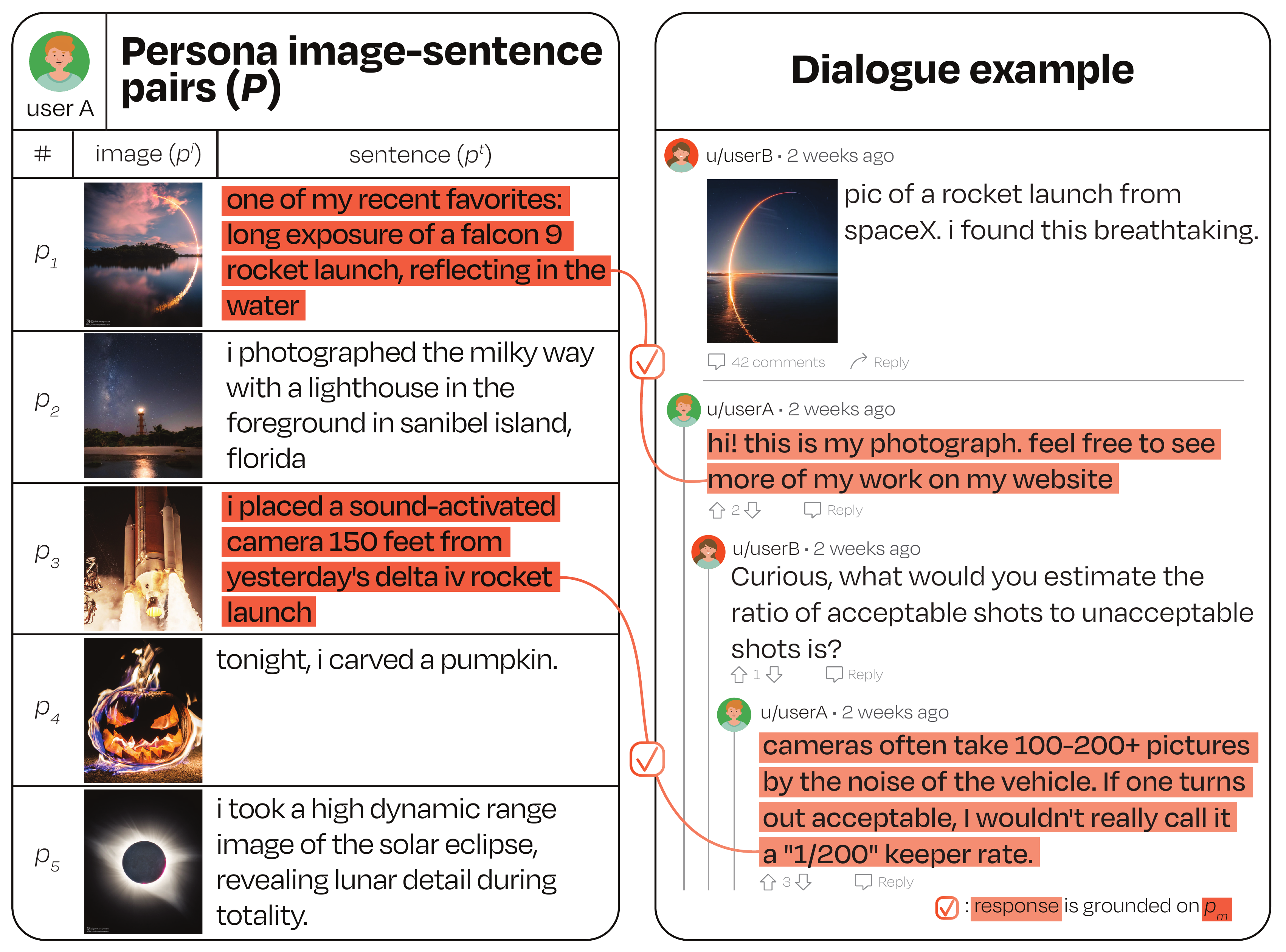}
    \caption{
        An example of \datasetName: a user A's persona (\ie five persona image-sentence pairs) in the left and a dialogue example
        in the right.
        Each persona element $p$ consists of a pair of an image $p^i$ and a sentence $p^t$.
        Each response from the user A in the dialogue example is grounded on a specific persona element $p_m$.
        Multimodal personas from \datasetName describe episodic memories of personal experiences (\eg favorite rockets and constellations) with visual details.
    }
    \label{fig:dataset_creation}
\end{center}
\end{figure}

However, existing studies restrict the role of persona only to personal facts~\citep{Zhang:2018:ACL} or personalities~\citep{Li:2020:AAAI}, while it should be explored in multi-faceted ways~\citep{Moore:2017:PS}.
More than factual information, episodic memory~\citep{Tulving:1972}, which is the memory of everyday events or personal experiences connected to the self and autonoetic consciousness~\citep{Tulving:2002:ARP,Conway:2005:JML}, should be included in persona component.
\citet{Wilson:2003:Memory} further supports this assertion by arguing that episodic memory plays a significant role in shaping personal identity, which in turn can influence one's persona.
Since episodic memories are often represented in the form of visual images or history scenes~\citep{Rubin:2003:MC,Conway:2009:Neuropsychologia}, we propose to study the \textit{multimodal persona},
which consists of a set of image-sentence pairs describing memorable moments as shown in Figure~\ref{fig:dataset_creation}.
Furthermore, visual information can complement textual information, which often lacks an explicit description of appearance or measurable quantities~\citep{Jin:2022:ACL,Zhang:2022:NAACL}.

In this work, we contribute to the persona-based dialogue research in two important ways.
First, we introduce a new multimodally personalized dialogue dataset named \textbf{M}ultimodal \textbf{P}ersona Chat (\datasetName), where personas reveal speakers' episodic-memories using both text and images.
To the best of our knowledge, \datasetName is the first dataset that
supports multimodal persona in dialogue.
To collect episodic-memory-based multimodal personas, we source users' posts from social media Reddit.
We carefully design a pipeline to curate multimodal conversation data that are well-grounded on multimodal personas\footnote{Note that our dataset pipelining approach is not restricted to Reddit and can be extended to other sources such as Twitter, Instagram, and more.}.

Second, based on \datasetName, we propose three retrieval-based dialogue tasks as benchmarks for multimodal persona-grounded dialogue understanding: next response prediction, grounding persona prediction, and speaker identification.
By incorporating our proposed multimodal persona, we observe statistically significant performance improvements across all tasks.

Consequently, our work illustrates the significance of multimodal persona in enhancing multimodal dialogue comprehension,
and our \datasetName provides a valuable resource for the research, given its well-grounded dialogues (especially responses) on multimodal personas.

\section{Related Work}
\label{sec:related_work}

\textbf{Personalized dialogue.}
Personalized dialogue agents have exploited \textit{persona} in the form of
unstructured sentences~\citep{Zhang:2018:ACL,Zhong:2020:EMNLP}, structured key-value attributes~\citep{Qian:IJCAI:2018,Zheng:2019:arXiv}, and personality types~\citep{Mairesse:2007:ACL,Wen:2021:ACLF}.
Persona in these works reveals only personal facts (\eg age, gender, job, location, hobby) or personalities (\eg Big-Five, MBTI) in the textual format.
Instead, we focus on an episodic-memory-based persona describing diverse, memorable moments of personal experiences~\citep{Schacter:2009} using both sentences and images.

\textbf{Multimodal datasets.}
To fuse visual and textual modalities, various works have been conducted on building datasets of paired images and text
~\citep{Ordonez:2011:NeurIPS,Lin:2014:ECCV,Krishna:2017:IJCV,Sharma:2018:ACL,Shao:2019:ICCV,Kuznetsova:2020:IJCV}
and multimodal models~\citep{Lu:2019:NeurIPS, Li:2020:ECCV, Li:2021:NeurIPS}.
In these datasets, text tends to explicitly describe the paired images (\eg image captioning and visual question answering) in a short sentence.
On the other hand, \citet{Desai:2021:NeurIPS} released RedCaps, whose image-sentence pairs are sourced from social media Reddit and whose text captions are more conversational and diverse than existing datasets.
We use Reddit to source image-sentence pairs as multimodal persona, but we build a new multi-turn dialogue dataset, \datasetName,
to extend the role of persona to reflect episodic memories and further explore multimodal dialogue comprehension in personalized dialogue.

\textbf{Multimodal dialogue.}
Research on multimodal (or image-grounded) dialogue has focused on understanding images and utterances in a context-aware manner
~\citep{Mostafazadeh:2017:IJCNLP, Das:2017:CVPR, Shuster:2020:ACL_IMAGECHAT, Zheng:2021:LREC, Zang:2021:ACL, Lee:2021:ACL}.
Simple retrieval dialogue agents~\citep{Shuster:2020:ACL_IMAGECHAT, Lee:2021:ACL}, which fuse textual and visual features, have been used to produce image-grounded responses.
\datasetName also consists of images and dialogues, but we utilize multimodal persona to produce both image-grounded and persona-grounded responses.

\section{The \datasetName Dataset}
We collect a multimodal persona-grounded dialogue dataset named \datasetName (\textbf{M}ultimodal \textbf{P}ersona \textbf{Chat}).
The objective of \datasetName is to help a conversational agent utilize its episodic-memory-based persona, consisting of both linguistic and visual information, to produce persona-grounded responses.
To cover a wide range of episodic-memory-based multimodal persona, we source posts from social media Reddit.

However, dialogue with a multimodal persona introduces two new challenges.
First, it is harder to collect persona image-sentence pairs than to
collect personas sentences.
Second, it is also difficult to collect dialogue instances grounded on speakers' multimodal personas since each utterance should be grounded on not only persona sentences but also persona images,
which may require more fine-grained information with additional commonsense knowledge~\citep{Cui:2020:EMNLPF,Liu:2022:ACL}.
To overcome these challenges, we design the process of data construction as follows.

\subsection{Collecting Multimodal Persona}
\label{subsec:multimodal_persona_construction}

Following RedCaps~\citep{Desai:2021:NeurIPS}, we manually curate a set of subreddits with a high proportion of image posts, 
where images are photographed by Reddit users themselves, and post titles are related to the image content.
In total, we use 648 subreddits, whose full list can be found in Appendix~\ref{subsec:persona_subreddit_list}.
We then download all image posts from the selected subreddits.
We intend to define a user's multimodal persona as $m$ number of image-sentence pairs where $m$ is the number of the user's posts. 
Thus, we group the downloaded posts according to users, and transform each post into a pair of one image and one sentence using 
(1) a rule-based method and (2) a model-based method as follows.

\textbf{Rule-based lexical method.}
We use the post title as the persona sentence.
If the title consists of multiple sentences, we select only the first one as done in \citet{Mazare:2018:EMNLP}.
We then retain the sentences that satisfy all the following rules: (1) each sentence must contain between 4 and 20 words, (2) it contains either the word \textit{I} or \textit{my}, and it consists of (3) at least one verb, (4) at least one noun or adjective, and (5) at least one content word.
With this method, we improve the fluency and expressiveness of the persona sentences.

\textbf{Model-based semantic method.}
After obtaining image-sentence pairs,
we ensure that the image is semantically relevant to its paired sentence.
We leverage the pretrained CLIP-ViT-B/32~\citep{Radford:2021:ICML}  to calculate semantic similarity between the image and the sentence, which is widely used in past research~\citep{Hessel:2021:EMNLP,Cho:2022:NAACLF,Frans:2022:NeurIPS}.
Then, we ignore the pair with a cosine similarity less than 0.

Finally, we follow \citet{Desai:2021:NeurIPS} to avoid potential ethical risks of curating Internet-scale image datasets.
See Appendix \ref{ethics_appendix} for the details of our ethical considerations.
As a result, about 10\% of downloaded posts are used to make multimodal personas, and the others can be exploited for dialogue data.  

\subsection{Collecting Dialogues}
\label{subsec:dialogue_construction}

Once we obtain a set of users' multimodal personas, we collect dialogue data where the users participate in the conversation.
Discussions on Reddit consist of \textit{threads}, each with one post and multiple comments, as shown in Figure~\ref{fig:dataset_creation}.
From the curated subreddits in Appendix~\ref{subsec:dialog_subreddit_list}, we collect threads containing the comments the users wrote with multimodal persona.
We exclude the threads used to make multimodal personas in \S~\ref{subsec:multimodal_persona_construction} to ensure that the source of persona is disjoint with that of conversation.
We iteratively trace the parent comment nodes in threads until the root node appears, finding the post and all its comments before the persona user's comment that constitutes a single conversation data.
Therefore, in each dialogue data, the last utterance spoken by the persona user becomes the \textit{response},
all previous comments and the image post become the \textit{context}.
We set the maximum number of turns in the context to 20.

We filter out dialogues where a user's response is posted earlier than the user's persona posts since the episodic-memory persona should chronologically precede the user's response.
We additionally filter dialogues as explained in Appendix~\ref{subsec:dialog_filtering}.

\subsection{Grounding Persona on Dialogues}
\label{subsec:multimodal_entailment}

To ensure persona-consistency, the user's response in dialogue should be well grounded on his or her multimodal persona.
Otherwise, it is impossible for an algorithm (or even a human) to correctly predict the response based on the persona, which may undermine the usefulness of our dataset.

We automatically filter out the conversations whose responses have no persona-related information by employing (1) heuristic rules and (2) pretrained models~\citep{Reimers:2019:EMNLP,Radford:2021:ICML}; see Appendix~\ref{subsec:filtering_rules_me} for details.

Despite the effectiveness of the automatic filtering process,
we empirically find that some responses are still not grounded on persona
since the pretrained models used for automatic filtering are not perfect.
According to~\citet{Welleck:2019:ACL}, identifying an utterance grounded on (\ie consistent with) a persona sentence can be reduced to a natural language inference (NLI) task.
Thus, we conduct additional human NLI annotation to make sure that the user's response is grounded on the multimodal persona.

In our NLI setting,
the premise $p=(p^i,p^t)$ is a persona image-sentence pair among the speaker's multimodal persona set $P=\{p_1,...,p_m\}$, and
the hypothesis $r$ is the response in conversation from the same speaker.
The goal is to perform a binary classification for a pair $(r,p)$:
(1) \entailmentLabelEntailed if there is enough evidence in $p=(p^i,p^t)$ to conclude that $r$ is most likely true.
(2) \entailmentLabelNotEntailed if
(i) there is enough evidence in $p$ to conclude that $r$ is most likely false, or
(ii) there is not enough evidence in $p$ to draw a conclusion about $r$.

We annotate entailment labels from human workers via Amazon Mechanical Turk (Mturk).
To reduce the label costs, we only collect entailment labels for at most two persona elements (among $m$ elements) per response $r$.
See Appendix~\ref{subsubsec:persona_response_pair_selection} on how to select the two persona elements.

Given a context $c=(c^t,c^i)$, response $r$ and a persona image-sentence pair $p$, we ask three annotators to categorize a pair $(r,p)$ into the two classes.
Following previous works~\citep{Bowman:2015:EMNLP,Xie:2019:arXiv}, we finalize labels according to the majority vote criterion (at least 2 out of 3).
As a result, we obtain the labels for 16,327 pairs from human workers, and 50.4\% of them are finally labeled as \entailmentLabelEntailed.
We defer the annotations’ details to Appendix~\ref{subsubsec:quality_control}.
The inter-annotator agreement for entailment labels is measured using Krippendorff’s $\alpha$ \citep{Krippendorff:2011:arXiv}. It is 0.47, implying a good agreement despite the difficulty of the task~\citep{Chen:2020:ACL,Zhang:2021:NAACL}.

\subsection{Final Multi-turn Dialogue Data}
\label{subsec:merging_into_final_multiturn_dialogues}
In summary, one dialogue consists of the \textit{response} as the last utterance spoken by the persona speaker 
and the \textit{context} as all prior utterances from the Reddit post.
We then construct a \textit{multi-turn dialogue} by merging the dialogues sharing common threads (\ie multiple responses by persona users exist in a single dialogue).
Finally, we have 7,898 multi-turn dialogue data whose responses are \entailmentLabelEntailed with (or grounded on) the persona (\ie at least one persona element-response pair is labeled as \entailmentLabelEntailed).
Also, we add a similar amount of dialogue data whose responses are grounded on no persona element,
since the dataset should be able to evaluate whether the method can correctly identify \textit{no grounding}.
It also follows \textit{persona-sparse} real-world conversations~\citep{Zheng:2020:AAAI} that contain a limited amount of dialogues grounded on speakers' persona.
By randomly selecting 7,102 such dialogues, eventually, \datasetName consists of 15,000 multi-turn dialogues. 

\subsection{Analysis of \datasetName Compared to Other Persona-Based Dialogue Datasets}
\label{subsec:dataset_analysis}
The dataset consists of 15,000 multi-turn dialogues with 42,531 utterances by 25,877 users.
We divide \datasetName into train/valid/test split with 11,975/1,516/1,509 dialogues chronologically;
the test set is the most recent dialogues so that they are disjoint with existing Reddit-sourced datasets.

\textbf{Statistics and properties.}
Table~\ref{tab:dataset_stats} compares \datasetName with other persona-based dialogue datasets.
Only \datasetName uses images for persona, and describes episodic-memory-based persona beyonds fact, thought, or personality.
Moreover, \datasetName provides additional persona entailment labels that indicate whether a response is grounded on a given image-sentence persona. 

{\renewcommand{\arraystretch}{1.0}
    \begin{table}[t!] \begin{center}
    \begin{adjustbox}{width=1.0\columnwidth}
    \begin{tabular}{lccccc}
        \toprule
        \makecell[l]{Dataset}                                & \makecell{\#Dialog}   & \makecell{Data\\source}   & \makecell{Persona\\type}  & \makecell{Persona\\modality} & \makecell{Entailment\\label}   \\
        \midrule
        \makecell[l]{LIGHT}             & 11K                   & CS                        & Fact                      & T                            & No                                            \\
        \makecell[l]{PD}                & 20.8M                 & Weibo                     & Fact                      & T                            & No                                            \\
        \makecell[l]{PEC}               & 355K                  & Reddit                    & Thought                   & T                            & No                                            \\
        \makecell[l]{PELD}              & 6.5K                  & TV shows                  & Personality               & T                            & No                                            \\
        \midrule
        \makecell[l]{PersonaChat}       & 13K                   & CS                        & Fact                      & T                            & Post-Hoc$^*$                                  \\
        \makecell[l]{FoCus}             & 14K                   & CS                        & Fact                      & T                            & Yes                                           \\
        \midrule
        \makecell[l]{\datasetName}      & 15K                   & Reddit                    & \makecell{Episodic\\memory}           & V,T                          & Yes                                           \\
        \bottomrule
    \end{tabular}
    \end{adjustbox}
    \caption{
        Comparison of \datasetName with other persona-based dialogue datasets: LIGHT~\citep{Urbanek:2019:EMNLP}, PD~\citep{Zheng:2019:arXiv}, PEC~\citep{Zhong:2020:EMNLP}, PELD~\citep{Wen:2021:ACLF}, PersonaChat~\citep{Zhang:2018:ACL} and FoCus~\citep{Jang:2022:AAAI}.
        CS indicates that crowd-sourced annotators write the dialogues and persona sentences.
        V and T denote visual and textual modality.
        $^*$The persona entailment labels of PersonaChat are collected later by another work \citep{Welleck:2019:ACL}.
    }
    \label{tab:dataset_stats}
\end{center}\end{table}}

\begin{figure}[t]
\begin{center}
    \includegraphics[width=\columnwidth]{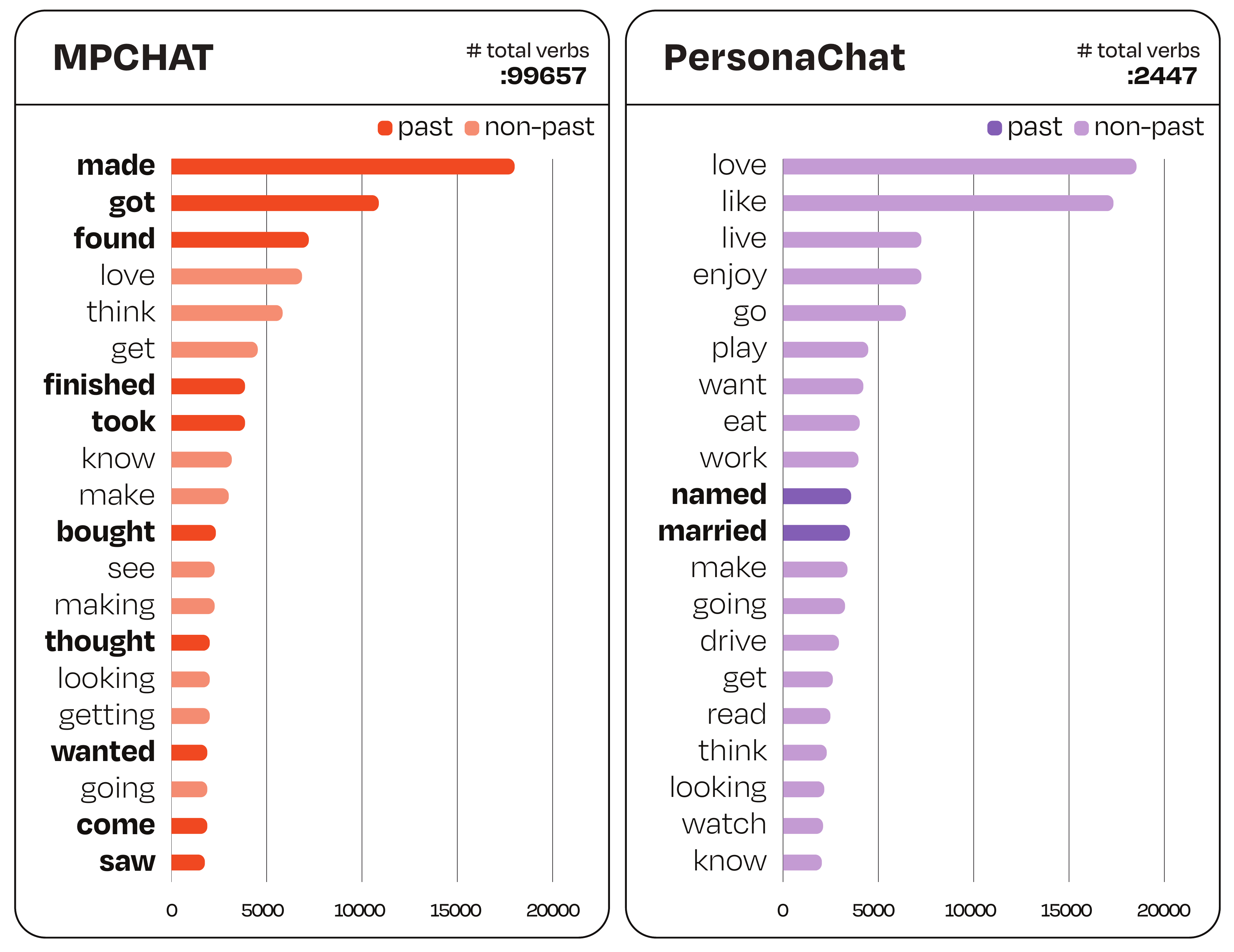}
    \caption{
        Comparison of the top-20 verbs between \datasetName and PersonaChat.}
    \label{fig:persona_frequency_analysis_verb}
\end{center}
\end{figure}

\textbf{Frequent verbs in personas.}
Figure~\ref{fig:persona_frequency_analysis_verb} compares the top-20 frequent verbs in persona sentences from \datasetName and PersonaChat~\citep{Zhang:2018:ACL}.
Thanks to Reddit's abundant sources, the number of verbs from \datasetName is much larger than those from PersonaChat.
The persona sentences in our dataset also include past tense verbs such as \textit{made, found}, and \textit{finished}
while persona sentences in PersonaChat do not.
It is because our personas are based on episodic memory,
which is the collection of personal experiences or memorable moments at particular times.

\textbf{Lexical diversity of personas.}
Table~\ref{tab:linguistic_properties} compares the lexical diversity of persona sentences from \datasetName with those from PersonaChat~\citep{Zhang:2018:ACL} and PEC~\citep{Zhong:2020:EMNLP}.
We count the number of N-grams from the fixed number (\ie 6,737) of randomly sampled persona sentences from each dataset.
Then, we measure lexical diversity using three metrics: MTLD, HD-D~\citep{McCarthy:2010:BRM} and MATTR scores~\citep{Covington:2010:JQL}.
Surprisingly, persona sentences from \datasetName achieve the highest scores in all lexical diversity metrics.
This result is also caused by the different properties of persona sentences: specific personal experiences of episodic memory in \datasetName \textit{vs.}  permanent characteristics, repeated events, and emotions in PersonaChat and PEC.

{\renewcommand{\arraystretch}{1.0}
    \begin{table}[t!] \begin{center}
    \begin{adjustbox}{width=1.0\columnwidth}
    \begin{tabular}{lcccccc}
    \toprule
    Dataset                         & \# 2-grams    & \# 3-grams    & \# 4-grams    & MTLD      & MATTR     & HD-D \\
    \midrule
    PersonaChat                     & 15,263        & 27,631        & 36,063        & 78.08     & 0.7791    & 0.7945 \\
    PEC                             & 34,051        & 54,649        & 62,290        & 111.39    & 0.811     & 0.8315 \\
    \datasetName                    & \textbf{39,694}        & \textbf{60,199}        & \textbf{66,732}        & \textbf{171.91}    & \textbf{0.8534}    & \textbf{0.8674} \\
    \bottomrule
    \end{tabular}

    \end{adjustbox}
    \caption{
      Lexical diversity comparison in the three metrics of MTLD, MATTR and HD-D scores
      based on the number of \{2, 3, 4\}-grams in each dataset.
    }
    \label{tab:linguistic_properties}
\end{center}\end{table}}

We report more dataset analyses in Appendix \ref{subsec:futher_analyses_mpchat}.

\begin{figure}[t]
\begin{center}
    \includegraphics[width=\columnwidth]{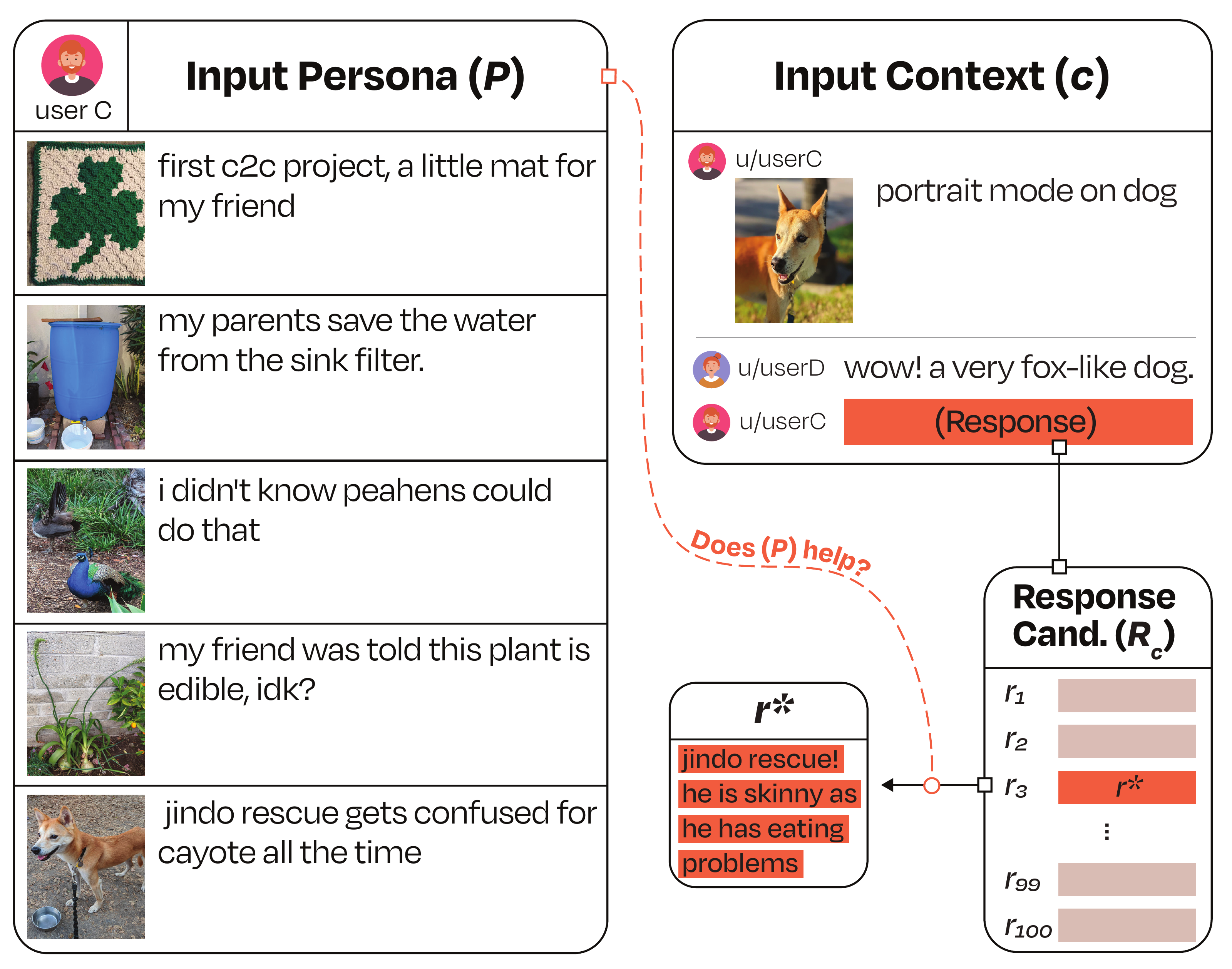}
    \caption{
        An example of the next response prediction.
    }
    \label{fig:tasks}
\end{center}
\end{figure}

\section{Task Definition}
\label{sec:task_definition}
As benchmark tasks for \datasetName, we consider three retrieval tasks as follows.
(1) The \textbf{next response prediction} task is to predict the next response given a context and the speaker's multimodal persona, which has been often regarded as a main task of persona-based dialogue~\citep{Humeau:2020:ICLR,Zhang:2018:ACL}.
(2) The \textbf{grounding persona prediction} task is to predict speaker's persona element,
either based on the dialogue context alone or based on both the dialogue context and the response.
This task is derived from and symmetrical to the next response prediction task.
Both the next response prediction and grounding persona prediction tasks are designed to ensure both multimodal context-awareness and multimodal persona-consistency.
(3) The \textbf{speaker identification} task is to identify the speaker participating in a dialogue given a context and a response,
which is crucial in personalized dialogues~\citep{Zhang:2018:ACL,Sang:2022:NAACL}.
In this task, we design it as a ranking problem, considering that \datasetName supports multi-party dialogues.
Furthermore, we expand the existing task into the multimodal domain.

Specifically, the dialogue dataset $D$ is a list of $N$ dialogues, each of which consist of $(c,r,P)$, 
where a context $c=(c^i,c^t)$ contains a context image $c^i$ and context text $c^t$ (\ie context utterances),
$r$ is a response to context $c$, and
a persona set $P=\{(p^i_1,p^t_1),...,(p^i_m,p^t_m)\}$ is a set of $m=5$ persona image-sentence pairs of the speaker who spoke the response $r$.
We below describe each task setting.

\textbf{Next response prediction.}
The goal of this task is to predict the next response $r^*$ based on $\Pr(r|c,P,R_{c})$, from a response candidate set $R_{c}=\{r_1, r_2, ..., r_{C_r}\}$,
as shown in Figure~\ref{fig:tasks}.
The response candidate set $R_{c}$ contains a correct response $r^*$ and $C_{r}-1$ randomly sampled test responses.

\textbf{Grounding persona prediction.}
This task aims at predicting the persona element $p^*$, which grounds $r$ (\ie labeled as \entailmentLabelEntailed in \S~\ref{subsec:multimodal_entailment}) based on $\Pr(p|c,r,\bar{P},P_{c})$ or $\Pr(p|c,\bar{P},P_{c})$.
$P_{c}=\{p_1,p_2,...,p_{C_p}\}$ is a persona (element) candidate set, which includes a correct persona element $p^*$ and $C_{p}-1$ randomly sampled persona elements from other speakers.  $\bar{P}$ is the speaker's remainder persona set, a set of $m-1$ persona image-sentence pairs in $P$ except $p^*$.
Note that we consider two cases of whether $r$ is given or not.
If $r$ is not given (\ie \texttt{no-response} case), then a model needs to retrieve the most likely persona element $p^*$ based on a given context $c$ and a remainder persona set $\bar{P}$ before producing a response $r$.
If $r$ is given (\ie \texttt{response} case), a model predicts $p^*$ that grounds $r$, which is much easier than the former case.

\textbf{Speaker identification.}
Finally, we predict the speaker (with his/her multimodal persona set) $P^*$ who spoke the response $r$ based on $\Pr(P|c,r,\mathbb{P}_{c})$,
from a speaker candidate set $\mathbb{P}_{c}=\{P_1,P_2,...,P_{C_P}\}$.
The speaker candidate set $\mathbb{P}_{c}$ includes a correct speaker $P^*$ and $C_{P}-1$ randomly sampled speakers.

Following \citet{Humeau:2020:ICLR,Zhong:2020:EMNLP,Shuster:2020:ACL_IMAGECHAT,Lee:2021:ACL}, we use Recall@1 and mean reciprocal rank (MRR) as evaluation metrics,
and set the number of retrieval candidates $C_r, C_p$, and $C_P$ to 100.

\section{Models}
\label{sec:models}
To solve the proposed retrieval-based dialogue tasks, we first define a set of unimodal encoders for the input of persona image and text $(P^i, P^t)$, context image and text $(c^i,c^t)$, and a response $r$.
We then construct multimodal persona-aware models by combining these modules based on input components for each task.
Note that we design our models to be simple and standard, to investigate the characteristics of our dataset. 

\textbf{Text encoder.}
We use a Transformer~\citep{Vaswani:2017:NeurIPS} as the text encoder for context text $c^t$, persona sentences $P^t$, and a response $r$.
We test two initialized weights of SBERT\footnote{\url{https://huggingface.co/sentence-transformers/multi-qa-distilbert-cos-v1}.}~\citep{Reimers:2019:EMNLP}
and the CLIP-ViT-B/32 text model~\citep{Radford:2021:ICML}.
For a persona input $P^t$, we encode the concatenation of $m$ persona sentences.
The representation of each text input ($h_{c^t},h_{P^t},h_{r}$) is obtained by the mean-pooled output of the entire sequence for SBERT or the hidden state of the first token \texttt{[CLS]} (for CLIP), followed by a linear layer.

\textbf{Image encoder.}
We encode a context image $c^i$ and a set of persona images $P^i$ using a single grid-based ViT-B/32~\citep{Dosovitskiy:2021:ICLR}
and CLIP-ViT-B/32 vision model~\citep{Radford:2021:ICML} due to its zero-shot ability.
We use the hidden states of the first patch of each image, followed by a linear layer, as a pooled representation following~\citet{Dosovitskiy:2021:ICLR}, which is mean-pooled to obtain a representation of persona images $h_{P^i}$.

\subsection{Models for Three Dialogue Tasks}

\begin{figure}[t]
\begin{center}
    \includegraphics[width=\columnwidth]{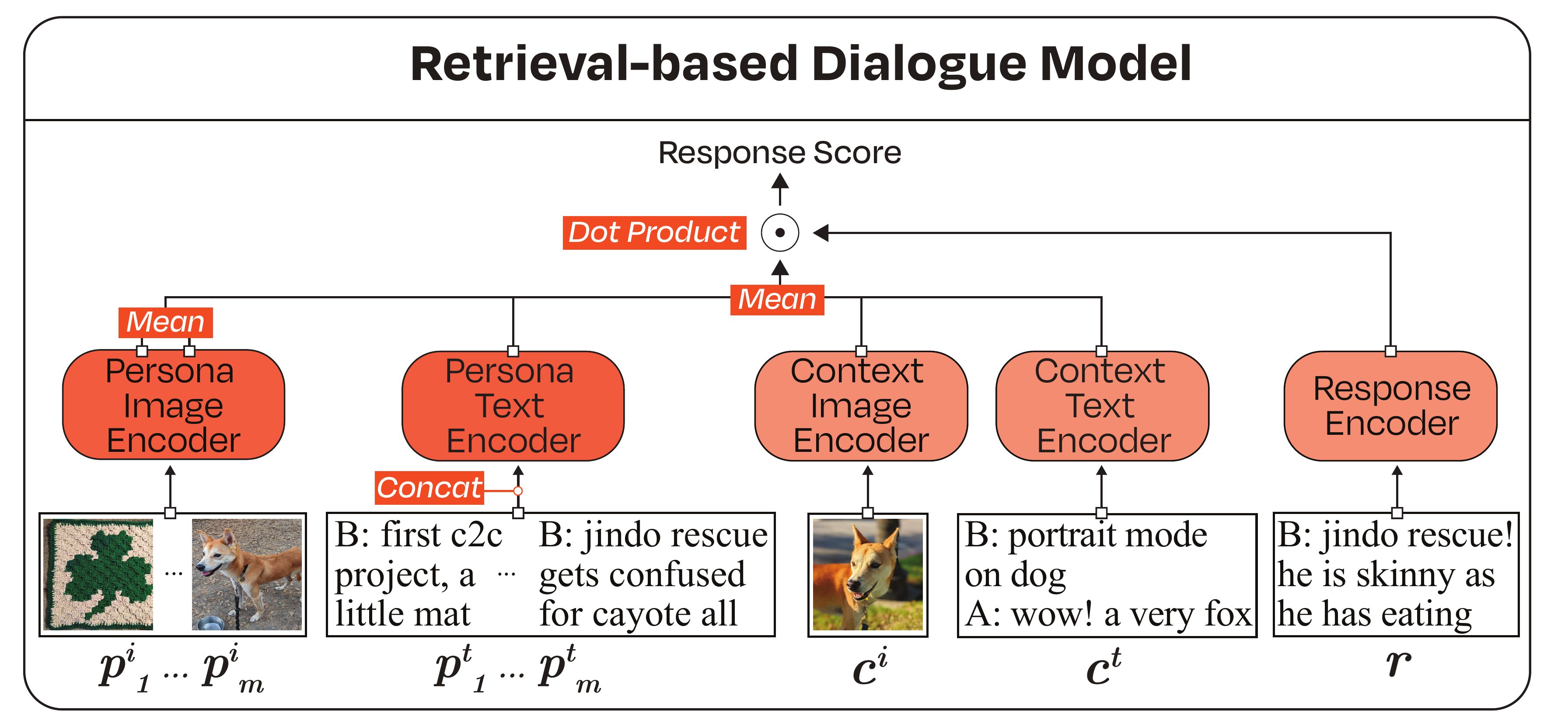}
    \caption{
        The architecture of retrieval-based model for the next response prediction task.
    }
    \label{fig:response_generation_model}
\end{center}
\end{figure}

Figure~\ref{fig:response_generation_model} shows our model for the next response prediction task, from which models for the two other tasks can be easily inferred.

\textbf{Next response prediction.}
After encoding each input separately, we first average $h_{P^i}$ and $h_{P^t}$ to produce the representation of a persona set $h_{P}$.
Then, we mean-pool $h_P,h_{c^t},h_{c^i}$ as the final representation $h_{out}$, which is used to compute the dot-product score for a response $r$ among candidate pool $R_{c}$ using $h_{out} \cdot h_r$.

\textbf{Grounding persona prediction.}
We first mean-pool $h_{\bar{P}^i}$ and $h_{\bar{P}^t}$ to obtain $h_{\bar{P}}$.
We then output $h_{out}$ by averaging all input embeddings of $h_{\bar{P}},h_{c^t},h_{c^i}$ for the \texttt{no-response} case and $h_r$ together for the \texttt{response} case.
Lastly, $h_{out}$ is used to compute the dot-product score for an image-sentence pair $p$ among candidate pool $P_{c}$ by $h_{out} \cdot h_p$,
where $h_p=\texttt{mean-pool}(h_{p^i},h_{p^t})$.

\textbf{Speaker identification.}
We mean-pool $h_{c^t},h_{c^i},h_r$ to produce $h_{out}$, which is used to compute the dot-product for a speaker's persona pairs $P=(P^i,P^t)$ among candidate pool $\mathbb{P}_{c}$ using $h_{out} \cdot h_P$,
where $h_P=\texttt{mean-pool}(h_{P^i},h_{P^t})$.

\subsection{Training and Inference}

According to encoder types, we test three conversation models: SBERT+ViT, SBERT+CLIP, and CLIP+CLIP (\ie original CLIP).
During training of all three tasks, we consider the other labels in each batch as negatives and train with a cross entropy loss over the matching scores as in \citet{Humeau:2020:ICLR}.
We do not update the parameters of image encoders (except CLIP+CLIP), which were common in previous studies~\citep{Shuster:2020:ACL_IMAGECHAT,Lee:2021:ACL}.
At the inference stage, each model selects the response that maximizes the dot-product score with the candidate set, such as 
$h_{out} \cdot h_{r_j}$ with $r_j \in R_{c}$ for next response prediction, the persona element $p_j \in P_{c}$ with $h_{out} \cdot h_{p_j}$ for persona prediction, and the speaker's persona $P_j \in \mathbb{P}_{c}$ with $h_{out} \cdot h_{P_j}$ for speaker identification.
We defer implementation details to Appendix~\ref{subsec:implementation_details}.

\section{Experiments}

The main goal of our experiments is to verify that multimodality from images and text indeed helps better understand persona-based dialogues, and our \datasetName is properly collected for this purpose.
Thus, we design our experiments as follows.
(1) Our models are rather simple and standard, as discussed in \S \ref{sec:models}.
(2) We compare our models that take advantage of full inputs with several baselines that use only parts of them.

\subsection{Next Response Prediction}
\label{subsec:experiment_mprs}
\textbf{Baselines.} We compare with the following baselines.
(1) Context text only ($c^t$):
This baseline outputs the matching score with the dot product between $h_{c^t}$ and $h_{r_j}$.
In addition, we add a simple information retrieval baseline, where the response candidates are arranged in the order of their weighted similarity (\ie TF-IDF score) to the context text $c^t$.
(2) Context image only ($c^i$):
It takes the dot product between $h_{c^i}$ and $h_{r_j}$ as the matching score.
(3) Context only ($c$): The matching score is the dot product between $h_c=\texttt{mean-pool}(h_{c^i},h_{c^t})$ and $h_{r_j}$.
(4) Context + persona sentences ($c,P^t$):
The matching score is the dot product between $h_{c;P^t}=\texttt{mean-pool}(h_{c^i},h_{c^t},h_{P^t})$ and $h_{r_j}$.
(5)
Context + persona images ($c,P^i$):
The matching score is the dot product between $h_{c;P^i}=\texttt{mean-pool}(h_{c^i},h_{c^t},h_{P^i})$ and $h_{r_j}$.

\textbf{Evaluation metrics.}
We evaluate the performance using Recall@1 and MRR metrics as described in \S~\ref{sec:task_definition}.
Statistical significance is computed using a two-sided t-test against the best competitor in all tasks,
including grounding persona prediction (\S~\ref{subsec:experiment_gpp}) and speaker identification (\S~\ref{subsec:experiment_si}).

\subsubsection{Results}
Table~\ref{tab:mprs} shows the results of next response prediction task.
We observe the following findings.

{\renewcommand{\arraystretch}{1.0}
    \begin{table}[t!]
      \begin{center}
       \small
    \begin{tabular}{llll}
        \toprule
        Model                                       & \makecell{R@1$\uparrow$}    & \makecell{MRR$\uparrow$}    \\
        \midrule
        \multicolumn{3}{l}{\textbf{Text Only} $(c^t)$}     \\
        IR Baseline     & 10.69              & 18.06             \\
        SBERT (zero-shot) & 35.67              & 45.75             \\
        SBERT           & 51.32$\pm$1.32              & 64.76$\pm$0.92             \\
        \midrule
        \multicolumn{3}{l}{\textbf{SBERT+ViT} (text + image encoder)}     \\
        $c$                  & 57.7$\pm$0.71   & 69.39$\pm$0.4             \\
        $c, P^i $          & 58.55$\pm$0.7   & 70.17$\pm$0.45             \\
        $c, P^t $          & 64.32$\pm$0.64    & 74.3$\pm$0.45             \\
        $c, P$ (Full)      & \textbf{65.29$\pm$0.66}$^{**}$    & \textbf{75.08$\pm$0.43}$^{**}$    \\
        \midrule
        \multicolumn{3}{l}{\textbf{SBERT+CLIP}}     \\
        $c$                   & 59.68$\pm$0.7              & 70.99$\pm$0.49             \\
        $c, P^i $           & 60.3$\pm$0.5             & 71.47$\pm$0.27             \\
        $c, P^t $           & 64.32$\pm$0.75             & 74.33$\pm$0.57             \\
        $c, P$ (Full)       & \textbf{65.43$\pm$0.42}$^{**}$    & \textbf{75.19$\pm$0.32}$^{**}$    \\
        \midrule
        \multicolumn{3}{l}{\textbf{CLIP+CLIP}}             \\
        $c^i$ (zero-shot)      & 39.38              & 54.06             \\
        $c^i$                  & 40.85$\pm$0.64              & 54.32$\pm$0.3             \\
        $c$                   & 69.11$\pm$0.74              & 78.22$\pm$0.49             \\
        $c, P^i $            & 69.87$\pm$0.4              & 78.85$\pm$0.27             \\
        $c, P^t $            & 72.13$\pm$0.61              & 80.72$\pm$0.38             \\
        $c, P$ (Full)       & \textbf{72.65$\pm$0.38}$^{*}$     & \textbf{81.12$\pm$0.26}$^{*}$    \\
        \bottomrule
    \end{tabular}
    \caption{Results of the next response prediction task. Symbols means $c^t$: context text, $c^i$: context image, $P^i$: persona images, and $P^t$: persona sentences.
        Also, $c= c^t \cup c^i $ and $P = P^i \cup P^t$.
        We report the average scores with standard deviations.
        Asterisks denote statistical significance of differences between \textit{full} model and its closest competitor  (*p < 0.05 and **p < 0.001).
    }
    \label{tab:mprs}
\end{center}\end{table}}

\textbf{Context image ($c^i$) helps response prediction.}
In all models, conditioning on the context image ($c^i$) significantly improves models to predict next response:
+7.34\% recall@1 score for SBERT+ViT model and +9.05\% recall@1 score for SBERT+CLIP model.
These performance gaps show that dialogues in \datasetName are well grounded on context images.
CLIP zero-shot model outperforms SBERT zero-shot model,
demonstrating CLIP's ability to retrieve the correct text response from the context image only.

\textbf{Persona images $P^i$ are important as well as persona sentences $P^t$.}
In all models, conditioning on persona images (\ie context + persona images) and on persona sentences (\ie context + persona sentences)
enhance next response prediction.
In addition, conditioning on persona sentences shows better performance than conditioning on persona images,
meaning that textual information in persona is more helpful than the image in persona to predict the textual response.

\textbf{Using both persona images $P^i$ and sentences $P^t$ achieves the best performance.}
In all models, using multimodal persona leads to the best Recall@1 and MRR scores.
It concludes that (1) \datasetName is well grounded on multimodal persona,
and (2) the persona image and sentence can complement each other to improve performance.

\subsection{Grounding Persona Prediction}
\label{subsec:experiment_gpp}
\textbf{Baselines.}
We use the following baselines.
We set the \texttt{no-response} as a default case.
(1) Context only ($c$):
The matching score is the dot product between $h_{p_j}$ and  $h_c=\texttt{mean-pool}(h_{c^i},h_{c^t})$ (or $h_{c;r}=\texttt{mean-pool}(h_{c^i},h_{c^t},h_r)$ for the \texttt{response} case).
(2) Context + remainder persona sentences ($c,\bar{P}^t$):
The matching score is the dot product between $h_{p_j}$ and $h_{c;\bar{P}^t}=\texttt{mean-pool}(h_{c^i},h_{c^t},h_{\bar{P}^t})$ (or $h_{c;r;\bar{P}^t}=\texttt{mean-pool}(h_{c^i},h_{c^t},h_r,h_{\bar{P}^t})$).
(3) Context + remainder persona images ($c,\bar{P}^i$):
The matching score is the dot product between $h_{p_j}$ and $h_{c;\bar{P}^i}=\texttt{mean-pool}(h_{c^i},h_{c^t},h_{\bar{P}^i})$ (or $h_{c;r;\bar{P}^i}=\texttt{mean-pool}(h_{c^i},h_{c^t},h_r,h_{\bar{P}^i})$).

\subsubsection{Results}
We present the results of grounding persona prediction in Table~\ref{tab:gpp} for the \texttt{no-response} as well as \texttt{response} cases.

\textbf{Providing response $r$ drastically improves performance.}
Compared to \texttt{no-response} case, results at \texttt{response} case indicate that
all models can predict the correct persona element based on the response with a 90\% chance or more,
meaning that persona entailment labels collected in \S~\ref{subsec:multimodal_entailment} are well annotated.

\textbf{Remainder persona images $\bar{P}^i$ provide visual clues.}
While not true for all cases, the results demonstrate that $\bar{P}^i$ improves models better than $\bar{P}^t$
in the following scenarios: CLIP+CLIP in both \texttt{no-response} and \texttt{response} cases, as well as CLIP+ViT in the \texttt{response} case.
Therefore, visual clues from $\bar{P}^i$ as well as textual clues from $\bar{P}^t$ are helpful in accurate persona prediction.

\textbf{Again, using both remainder persona images $\bar{P}^i$ and sentences $\bar{P}^t$ maximizes the performance.}
In both cases, models equipped with full inputs attain the best Recall@1 and MRR scores.
It verifies the usefulness of the multimodal remainder persona set $\bar{P}=(\bar{P}^i,\bar{P}^t)$.

{\renewcommand{\arraystretch}{1.0}
    \begin{table}[t!]
    \begin{center}
    \begin{adjustbox}{width=\columnwidth}
    \begin{tabular}{lllll}
        \toprule
        \multirow{2}{*}{Model} & \multicolumn{2}{l}{\makecell{\texttt{no-response}}} & \multicolumn{2}{l}{\makecell{\texttt{response} (+$r$)}} \\
        \cmidrule(r{0.3em}){2-3} \cmidrule(r{0.3em}){4-5}
         & \makecell{R@1$\uparrow$} & \makecell{MRR$\uparrow$} & \makecell{R@1$\uparrow$} & \makecell{MRR$\uparrow$} \\
        \midrule
        \multicolumn{5}{l}{\textbf{SBERT+ViT}}     \\
        $c$                   & 70.91$\pm$0.7             & 79.26$\pm$0.47                                       & 95.06$\pm$0.32             & 97.12$\pm$0.17             \\
        $c,\bar{P}^i$     & 70.7$\pm$0.9             & 79.17$\pm$0.57                                            & 95.16$\pm$0.55             & 97.21$\pm$0.29             \\
        $c,\bar{P}^t$     & 73.87$\pm$0.65             & 81.41$\pm$0.34                                          & 94.86$\pm$1.35             & 97.09$\pm$0.78             \\
        $c,\bar{P}$ (Full)  & \textbf{74.43$\pm$0.64}$^*$    & \textbf{82.05$\pm$0.39}$^{**}$                    & \textbf{95.75$\pm$0.53}$^{**}$    & \textbf{97.58$\pm$0.3}$^{**}$    \\
        \midrule
        \multicolumn{5}{l}{\textbf{SBERT+CLIP}}     \\
        $c$                   & 70.98$\pm$0.94             & 79.28$\pm$0.56                                           & 94.99$\pm$0.55             & 97.06$\pm$0.31             \\
        $c,\bar{P}^i$     & 70.63$\pm$1.03             & 79.22$\pm$0.71                                    & 94.91$\pm$0.44             & 97.04$\pm$0.24             \\
        $c,\bar{P}^t$     & 74.06$\pm$0.68             & 81.52$\pm$0.42                                    & 94.92$\pm$0.42             & 97.13$\pm$0.26             \\
        $c,\bar{P}$ (Full)  & \textbf{74.69$\pm$0.62}$^*$    & \textbf{82.24$\pm$0.41}$^{**}$              & \textbf{95.55$\pm$0.58}$^*$    & \textbf{97.48$\pm$0.32}$^{**}$    \\
        \midrule
        \multicolumn{5}{l}{\textbf{CLIP+CLIP}}     \\
        $c$                   & 78.85$\pm$1.04             & 85.96$\pm$0.67                          & 93.56$\pm$0.56             & 96.21$\pm$0.37             \\
        $c,\bar{P}^i$     & 82.02$\pm$0.89             & 88.31$\pm$0.58                             & 94.62$\pm$0.48             & 96.86$\pm$0.32             \\
        $c,\bar{P}^t$     & 80.69$\pm$0.8             & 87.28$\pm$0.55                              & 94.43$\pm$0.45             & 96.79$\pm$0.23             \\
        $c,\bar{P}$ (Full) & \textbf{82.32$\pm$0.75}    & \textbf{88.52$\pm$0.46}                   & \textbf{94.79$\pm$0.5}    & \textbf{96.94$\pm$0.28}    \\
        \bottomrule
    \end{tabular}
    \end{adjustbox}
    \caption{Results of the grounding persona prediction task in both \texttt{no-response} and \texttt{response} cases.
      Symbols means $c$: context text and image, $r$: response, $\bar{P}^i$: remainder persona images, $\bar{P}^t$: remainder persona sentences, and $\bar{P}=\bar{P}^i \cup \bar{P}^t$.
      Note that we include response $r$ as an additional input to the model only in the \texttt{response} case.
      We report the average scores with standard deviations.
      Asterisks denote statistical significance of differences between \textit{full} model and its closest competitor  (*p < 0.05 and **p < 0.001).
    }
    \label{tab:gpp}
\end{center}\end{table}}

\subsection{Speaker Identification}
\label{subsec:experiment_si}
\textbf{Baselines}.
(1) Text only dialogue ($c^t,r$) + speaker's persona sentences ($P_j^t$):
The matching score is the dot product between $h_{c^t;r}=\texttt{mean-pool}(h_{c^t},h_{r})$ and $h_{P_j^t}$.
(2) Dialogue ($c,r$) + speaker's persona sentences ($P_j^t$):
The matching score is the dot product between $h_{c;r}=\texttt{mean-pool}(h_{c^i},h_{c^t},h_{r})$ and $h_{P_j^t}$.
(3) Dialogue ($c,r$) + speaker's persona images ($P_j^i$):
The matching score is the dot product between $h_{c;r}=\texttt{mean-pool}(h_{c^i},h_{c^t},h_{r})$ and $h_{P_i^t}$.

\subsubsection{Results}
From Table~\ref{tab:si}, we can find several observations about the speaker identification task.

\textbf{Persona sentences $P_j^t$ are more important than persona images $P_j^i$.}
In all models, predicting the speaker based on his/her persona sentences $P_j^t$ outperforms that on persona images $P_i^t$.
It indicates that textual information plays a key role in retrieving the right speaker in this task.

\textbf{Using multimodal information $P_j$ still enhances speaker identification.}
In all models, identifying the speaker based on his/her persona image-sentence pairs $P_j=(P_j^i,P_j^t)$ shows the highest scores.
That is, persona images can complement persona sentences, showing the necessity of multimodal persona for the speaker identification task.

Furthermore, we present additional analyses that go beyond the main
experiments in Appendix \ref{sec:further_analyses_experiments}.

{\renewcommand{\arraystretch}{1.0}
    \begin{table}[t!]
    \begin{center}
    \small
    \begin{tabular}{llll}
        \toprule
        Model                                  & \makecell{R@1$\uparrow$}    & \makecell{MRR$\uparrow$}    \\
        \midrule
        \multicolumn{3}{l}{\textbf{Text Only} $(c^t,r, \mathbb{P}_{c}^{t})$}     \\
        SBERT                           & 56.47$\pm$0.58             & 67.92$\pm$0.52             \\
        \midrule
        \multicolumn{3}{l}{\textbf{SBERT+ViT}}     \\
        $c,r,\mathbb{P}_{c}^{i}$       & 19.56$\pm$0.64             & 35.84$\pm$0.45             \\
        $c,r,\mathbb{P}_{c}^{t}$       & 56.87$\pm$0.6             & 68.33$\pm$0.37             \\
        $c,r,\mathbb{P}_{c}$ (Full)          & \textbf{57.28$\pm$0.44}    & \textbf{68.86$\pm$0.3}$^{**}$    \\
        \midrule
        \multicolumn{3}{l}{\textbf{SBERT+CLIP}}     \\
        $c,r,\mathbb{P}_{c}^{i}$       & 25.71$\pm$0.49              & 42.47$\pm$0.34             \\
        $c,r,\mathbb{P}_{c}^{t}$       & 56.63$\pm$0.66             & 68.15$\pm$0.42             \\
        $c,r,\mathbb{P}_{c}$ (Full)          & \textbf{57.24$\pm$0.63}$^{*}$    & \textbf{68.69$\pm$0.39}$^{*}$             \\
        \midrule
        \multicolumn{3}{l}{\textbf{CLIP+CLIP}}     \\
        $c,r,\mathbb{P}_{c}^{i}$       & 44.27$\pm$0.66             & 59.04$\pm$0.35             \\
        $c,r,\mathbb{P}_{c}^{t}$       & 59.89$\pm$0.71              & 70.87$\pm$0.53             \\
        $c,r,\mathbb{P}_{c}$ (Full)          & \textbf{62.17$\pm$0.56}$^{**}$     & \textbf{73.08$\pm$0.35}$^{**}$             \\
        \bottomrule
    \end{tabular}
    \caption{Results of the speaker identification task.
        $\mathbb{P}_{c}=(\mathbb{P}_{c}^{i},\mathbb{P}_{c}^{i})$ is a
        speaker candidate set from which the speaker is retrieved, consisting of a set of speakers' persona images $\mathbb{P}_{c}^{i}$ and sentences $\mathbb{P}_{c}^{t}$. Symbols mean $c$: context text and image, and $r$: response.
        We report the average scores with standard deviations.
        Asterisks denote statistical significance of differences between \textit{full} model and its closest competitor  (*p < 0.05 and **p < 0.001).
    }
    \label{tab:si}
\end{center}\end{table}}

\subsection{Error Analysis}
We investigate error cases, specifically focusing on next response prediction and grounding persona prediction (\texttt{no-response}) tasks.
We analyze missed retrieved responses/persona and discuss factors related to multimodal comprehension and understanding of both dialogue context and persona information.

\subsubsection{Next Response Prediction}
We randomly selected 30 examples from the 629 incorrect predictions made by the CLIP+CLIP (with \textit{full} inputs) out of the test set.
Among them, we observed the following patterns in errors:

\textbf{Multimodal understanding.}
19 instances (63\%) failed in multimodal understanding,
indicating challenges in effectively leveraging both visual and textual information.
Specifically, 14 instances required multi-hop reasoning between the multimodal context ($c^i,c^t$) and multimodal persona components ($P^i,P^t$), such as cases involving visual coreference resolution.
Additionally, 5 instances solely relied on context comprehension ($c$ only) without considering persona information.

\textbf{Text understanding.}
9 instances (30\%) struggled with text understanding,
indicating persistent difficulties in comprehending complex textual clues.
Out of these instances, 7 required multi-hop reasoning between the context $c^t$ and persona $P^t$,
while 2 instances required context comprehension ($c^t$ only) without considering persona information.

\textbf{Task ambiguity.}
2 instances (7\%) failed due to the task ambiguity, where the next response $r^*$ is not the only response given context $c$ and a persona set $P$.

\subsubsection{Grounding Persona Prediction (\texttt{no-response})}
We randomly selected 30 examples from the 123 incorrect predictions made by the CLIP+CLIP (with \textit{full} inputs) out of the test set, and identified the following error patterns:

\textbf{Multimodal understanding.}
Among the instances, 17 (57\%) failed in multimodal understanding.
15 instances required multi-hop reasoning between the multimodal context ($c^i,c^t$) and multimodal persona components ($\bar{P^i},\bar{P^t}$),
while 2 instances required persona-consistency comprehension ($\bar{P}$ only) without context information.

\textbf{Text understanding.}
9 instances (30\%) failed in text understanding.
Out of these, 7 required multi-hop reasoning between the context $c^t$ and persona $P^t$.
2 instances required persona-consistency comprehension ($\bar{P^t}$ only) without considering context information.

\textbf{Task ambiguity.}
In 4 instances (13\%), errors were caused by task ambiguity, where the persona element $p^*$ is not the only answer given context $c$ and a remainder persona set $\bar{P}$.

These results highlight the challenges in effectively leveraging multimodal information and
emphasize that understanding both multimodal context and multimodal persona poses a greater challenge for dialogue models compared to understanding context or persona alone.

\section{Conclusion}
We studied episodic-memory-based \textit{multimodal} persona-grounded
dialogue, and introduced \datasetName as the first multimodal persona-grounded multi-turn
dialogue dataset.
We proposed three retrieval-based dialogue tasks to evaluate the effectiveness of multimodal persona.
With the help of multimodal persona, all of the proposed models exhibited better dialogue comprehension abilities.
Our empirical results showed that dialogues (especially responses) in \datasetName are well grounded on multimodal personas as intended.
One interesting future work would be to expand \datasetName in both the size (\eg scaling up the number of dialogues and personas) and the scope (\eg adding audio/video modality).

\section*{Limitations}
\label{limitations}
Since \datasetName sources the data from Reddit, it has the limitation that it may not be representative of the general population.
First, all subreddits of \datasetName are primarily written in English, and a significant percentage of Reddit users are from English-speaking countries.
The four countries with the highest desktop traffic on Reddit are the US, UK, New Zealand, and Australia, accounting for 66\% of the total user \citep{limitation:statistica}.
Moreover, compared to the average US population, \citet{limitation:pew} reported that Reddit users
are more likely to be male (67\% vs. 49\%), young (64\% 18-29 years old vs. 22\%), college-educated (42\% vs. 28\%), and politically liberal (43\% vs. 24\%). 
Therefore, \datasetName may reflect such somewhat narrow interests, and the demographic group represented by our model may be biased toward personal conversations suitable for it.

\section*{Ethics Statement}
\label{ethics}
We put much effort into ensuring that our \datasetName dataset includes no personal identifying information (PII):
we only picked subreddits that were not aimed at people and filtered out faces, license plates, and email addresses. 
Also, we only selected subreddits without 18+ tags and filtered NSFW images, offensive words, etc.
Note that we \textbf{manually filtered out} all images containing PII or NSFW content before publicly releasing \datasetName.
Human annotators earned an average wage of \$16 per hour, above the minimum wage in their areas. 
We abided by the Reddit API Terms of Use and also informed our annotators about this. 
Finally, we specified all licenses of scientific artifacts and will include them when distributing our data. 
See Appendix \ref{ethics_appendix} and \ref{licenses} for the details.

However, potential risks still remain in our data. 
As mentioned in Limitations~\ref{limitations} and Appendix~\ref{subsubsec:quality_control}, authors and annotators of \datasetName are primarily in the US, UK, New Zealand, and Australia.
These demographic and geographic biases mean that \datasetName may not equally represent all groups. 
Meanwhile, \citet{Wang:2021:arXiv, Lee:2022:arXiv} reported that preprocessing data with CLIP can cause gender-bias issues. 
We use CLIP to measure image-text similarity in the pre-processing for
data collection, so this problem may exist in our dataset.

Users of our dataset should be aware of these risks. 
To comply with the Reddit API Terms of Use and to protect the privacy of Reddit users, commercial and for-profit use of our data is limited.
It must be available for academic purposes only.

\section*{Acknowledgements}
First of all, we thank all our workers on MTurk for their dedication and enormous contribution to constructing \datasetName through this project.
We would also like to thank Hyunwoo Kim, Jiwan Chung, Soochan Lee, Jinseo Jeong, Insu Jeon, Jaekyeom Kim, Euihyun Tae, and the anonymous reviewers for their valuable comments.
This work was supported by Samsung Research Funding Center of Samsung Electronics (No. SRFCIT2101-01) and Institute of Information \& communications Technology Planning \& Evaluation (IITP) grant funded by the Korea government (MSIT) 
(No.2021-0-01343, Artificial Intelligence Graduate School Program for Seoul National University, and  No.2022-0-00156, Fundamental research on continual meta-learning for quality enhancement of casual videos and their 3D metaverse transformation).
Gunhee Kim is the corresponding author.

\bibliography{acl2023_dialog.bbl}
\bibliographystyle{acl_natbib}

\clearpage


\appendix

\section*{Appendix}
\label{sec:appendix}

\section{More details on Dataset Collection}

\subsection{Filtering Dialogue Data}
\label{subsec:dialog_filtering}
We filter Reddit conversation data to ensure that (1) each post is between 2 and 100 words, and (2) each comment is between 2 and 60 words\footnote{This is because posts are usually longer than comments.}.
We remove dialogues whose images contain potential ethical risks; see
Appendix \ref{ethics_appendix} for the ethical considerations in detail.
We automatically filter out whose utterances contain words or phrases from a blocklist\footnote{\url{https://github.com/rominf/profanity-filter}}
to prevent models from training offensive expressions.
Also, we ignore dialogues that are written earlier than the user's multimodal persona.
This is because a multimodal persona represents episodic memory in history,
and thus predicting responses in conversations that precede the persona may not be reasonable.
Finally, we lowercase all text and remove emojis, special symbols, URLs, and email IDs (including ``@'') from each sentence.

\subsection{Automatic Filtering of Persona Irrelevant Conversation}
\label{subsec:filtering_rules_me}
Given a dialogue context that consists of image $c^i$ and text $c^t$ parts and a response $r$,
and a set of persona image-sentence pairs $P=\{(p_1^i, p_1^t),...,(p_j^i,p_j^t),...,(p_m^i, p_m^t)\}$ of the speaker who wrote $r$,
we filter the conversation as follows.

We first filter out the conversation if the length of the response ($r$) is shorter than five words
because short responses usually do not contain persona-related information.

Next, we keep the conversation if any persona element $(p_j^i,p_j^t)$ in $P$ is related to the response $r$ as follows:
we measure the text similarity (\ie cosine similarity) score between the response and the persona sentence $sim_{SBERT}(r,p_j^t)$
and again measure the text similarity score between the context text and the persona sentence $sim_{SBERT}(c^t,p_j^t)$
by employing a Sentence BERT (or SBERT) model\footnote{\url{https://huggingface.co/sentence-transformers/all-MiniLM-L6-v2}}~\citep{Reimers:2019:EMNLP}.
After manually checking some data instances, we set a threshold of 0.5 to filter out instances in which $r$ is not related to $p_j^t$.
That is, if both $sim_{SBERT}(r,p_j^t)$ and $sim_{SBERT}(c^t,p_j^t)$ are below the threshold, we filter out the persona element.

We also measure the image-text similarity (\ie cosine similarity) between the response and the persona image $sim_{CLIP}(r,p_j^i)$
and again measure the similarity between the context text and the persona image $sim_{CLIP}(c^t, p_j^i)$
by employing a CLIP-ViT-B/32 model~\citep{Radford:2021:ICML}.
In this case, we set a threshold of 0 to filter out no persona-related conversations,
and if either $sim_{CLIP}(r, p_j^i)$ or $sim_{CLIP}(c^t, p_j^i)$ is below the threshold, we filter out the persona element.

After all, we keep the conversation if any of the persona elements are unfiltered.

\subsection{Details on Persona Entailment Labeling}
\subsubsection{Two-Class Persona Entailment}
\label{subsubsec:persona_entailment_binary_classification}

Unlike previous works \citep{Williams:2018:NAACL,Welleck:2019:ACL} that use 3-way labels of $\{$entailment, contradiction, neutral$\}$,
we modify it to 2-way labels of $\{$\entailmentLabelEntailed, \entailmentLabelNotEntailed$\}$ since we are interested in the detection of persona-response grounding.
Also, we find that the same speaker is unlikely to post contradictory sentences (or images),
leading to merging \textit{contradicted} and \textit{neutral} labels into \entailmentLabelNotEntailed label.

\subsubsection{Persona Selection for Entailment Labeling}
\label{subsubsec:persona_response_pair_selection}
Given a dialogue with a context image $c^i$, context text $c^t$ and a response $r$,
and a set of persona elements $P=\{(p_1^i, p_1^t),...,(p_j^i,p_j^t),...,(p_m^i, p_m^t)\}$ of the speaker who wrote $r$,
we select at most two persona elements per response $r$ as follows.
First, we apply the same method as in Appendix~\ref{subsec:filtering_rules_me} to filter out no persona-related response.
We drop the whole dialogue and do not select any persona element if all elements are filtered out.
If only one persona element is survived, then we select it.
If multiple persona elements are survived,
we select at most two persona elements based on text similarity scores:
(1) an element with the best $sim_{SBERT}(r,p_j^t)$ score and (2) one with the best score of the sum of $sim_{SBERT}(r,p_j^t)+sim_{SBERT}(c^t,p_j^t)$.
Then the persona elements selection is over, and the remaining data (\ie a set of at most two persona element-dialogue pairs) moves on to the next step: human annotations for the persona entailment labeling task.

\subsubsection{UI design for Mturk}
\begin{figure*}[t]
\begin{center}
    \includegraphics[width=\linewidth]{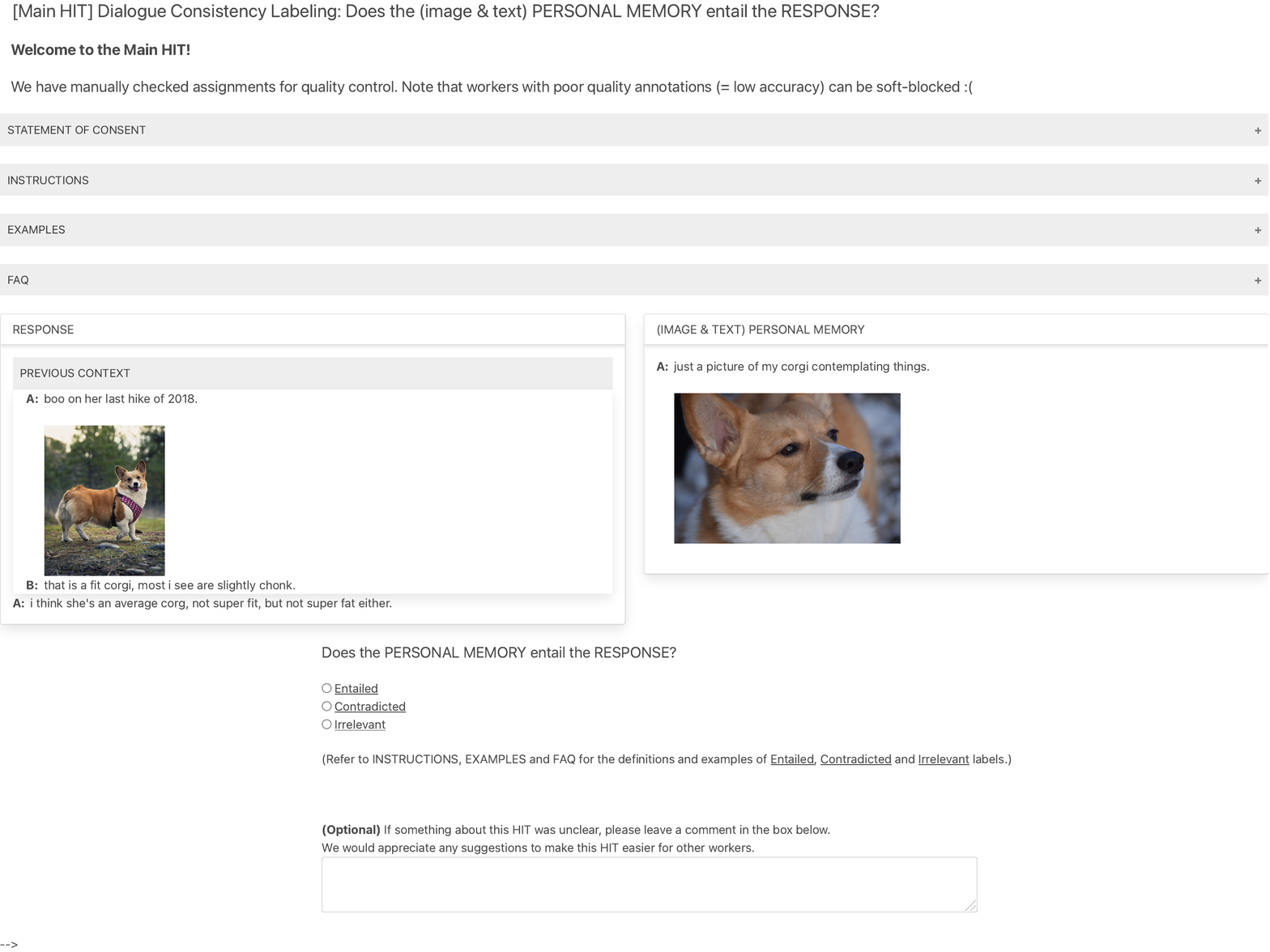}
    \caption{
     The UI design of Amazon Mechanical Turk to collect human annotations for persona entailment labels.
     }
    \label{fig:amt_meid}
\end{center}
\end{figure*}

\begin{figure*}
    \begin{center}
        \includegraphics[width=\linewidth]{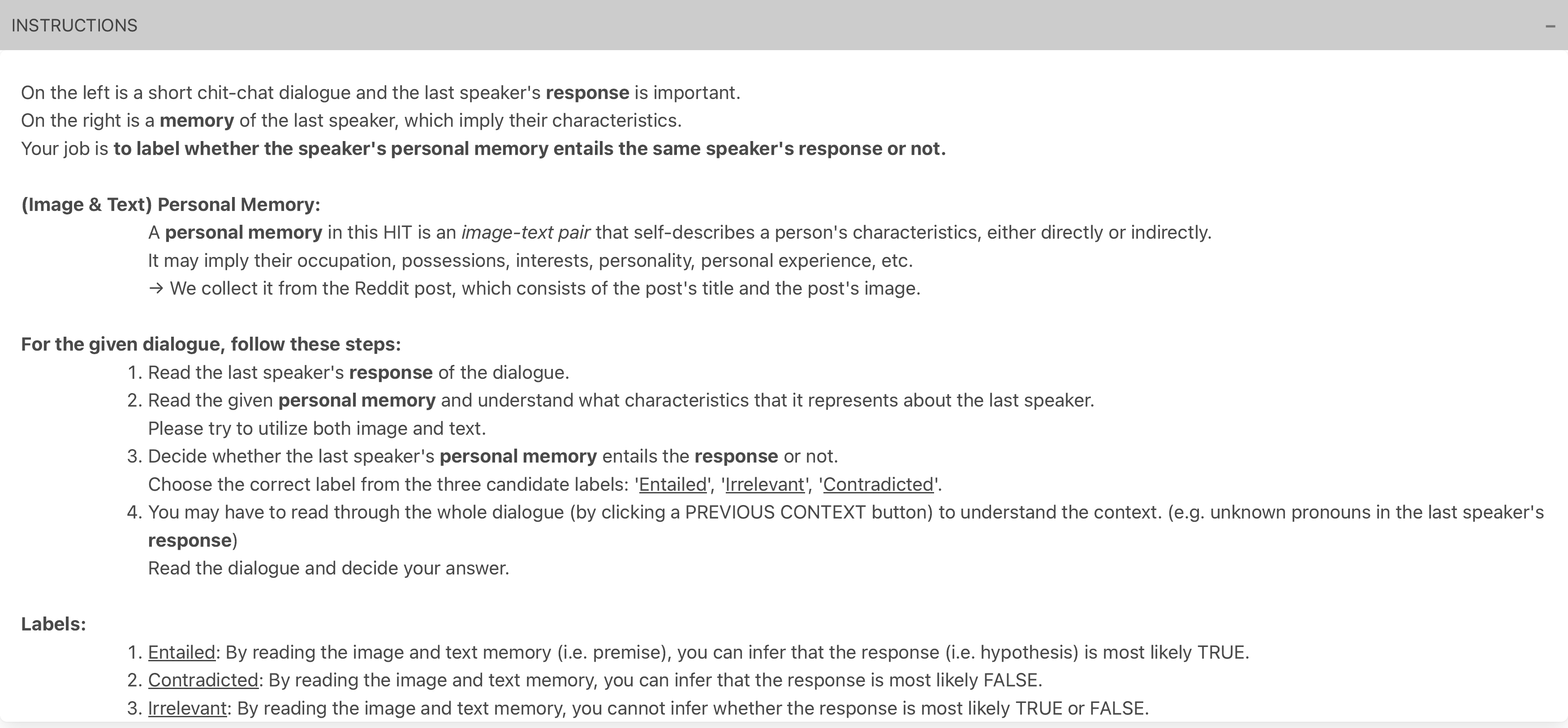}
        \caption{
         The instructions in the UI design of Amazon Mechanical Turk to collect human annotations for persona entailment labels.
         }
        \label{fig:amt_instructions}
    \end{center}
    \end{figure*}

Figure~\ref{fig:amt_meid} and Figure~\ref{fig:amt_instructions} show the annotation page for annotators labeling persona entailment labels. 
Note that we provide 3-way labels among \textit{entailed}, \textit{contradicted}, and \textit{irrelevant} (\ie \textit{neutral}),
and then reduce them to 2-way labels by merging \textit{contradicted} and \textit{irrelevant} into \entailmentLabelNotEntailed, while maintaining \textit{entailed} label as \entailmentLabelEntailed.

\subsubsection{Quality Control for Human Annotators}
\label{subsubsec:quality_control}
We only allow annotators located at one of [AU, CA, NZ, US, GB].
We use a qualification test to discern annotators who do not fully understand the task
(\eg only selecting \entailmentLabelNotEntailed regardless of the problem, or selecting \entailmentLabelEntailed just because $r$ and $p^t$ seem to be lexically similar).
Based on submitted answers in the qualification, we manually approve workers if they earn an acceptable score.
We periodically block malicious annotators to maintain high approval rates, while providing a reasonable bonus to benevolent workers.
Moreover, we steadily profile workers whose accuracy is lower than the average and re-educate them by showing examples with detailed explanations.
As a result, a total of 65 workers participated in the annotation process.

\subsubsection{Persona Entailment Label Distribution}
\label{subsubsec:entailment_label_distribution}
Table~\ref{tab:meid_stats} indicates the statistics of persona entailment labels.
Only 50.4\% of persona-response pairs are labeled as \entailmentLabelEntailed,
meaning that the proposed automatic filtering method alone is not enough,
while human-annotated persona entailment labels improve persona-groundedness of \datasetName.

{\renewcommand{\arraystretch}{1.0}
    \begin{table}[t!] \begin{center}
    \small
    \begin{tabular}{cccc}
        \toprule
                \makecell{\# Persona-\\response pairs}    & \makecell{\# Class}     & \makecell{\# Annotations\\per instance}    & \makecell{\# Labels \\(E/NE)}  \\
        \midrule
                    16,327            & 2             & 3              & 8,235/8,092                \\
        \bottomrule
    \end{tabular}
    \caption{Statistics of persona entailment labels.
        E: \entailmentLabelEntailed, and NE: \entailmentLabelNotEntailed.
    }
    \label{tab:meid_stats}
\end{center}\end{table}}

\subsection{Ethical Considerations in Data Collection}
\label{ethics_appendix}
In our data collection, we follow the overall ethical considerations proposed by
RedCaps~\citep{Desai:2021:NeurIPS} to align with the Reddit API terms of use and avoid violating ethical principles. 
We perform additional efforts to protect user privacy, such as license plate detection.

\textbf{Privacy}.
The foremost consideration for us is to protect the privacy of Reddit users.
Although \datasetName gathers `persona' data of each speaker in the dialogues, we try not to involve private information. The details are as follows. 
\begin{enumerate}
    \item We manually select the subreddits that are not focused on describing people. The resulting subreddits are mainly about general photography, animals, plants, objects, food, scenery, or activities.
    \item We perform automatic data filtering with RetinaFace~\citep{Deng:2019:arXiv} to remove any image with a human face with confidence $\geq$ 0.9.
    \item We automatically detect license plates using an open source detector\footnote{\url{https://github.com/ThorPham/License-plate-detection}} and filter out corresponding images with confidence $\geq$ 0.5.
    \item From the dialogue text, we delete any URL and email address (detected by ``@'') to avoid mentioning any explicit references to SNS IDs or email addresses.
\end{enumerate}

\textbf{Harmful contents}.
We also filter out offensive, insulting, or threatening content with the following steps:
\begin{enumerate}
    \item We manually select only non-NSFW(\ie not safe for work) subreddits.
    \item Within the curated subreddits, we do not include posts with over 18 tags.
    \item We perform automatic data filtering through InceptionV3~\citep{Szegedy:2016:CVPR} 
        from an open source model\footnote{\url{https://github.com/GantMan/nsfw_model}} with confidence $\geq$ 0.031.
        All data instances that include images classified into \textit{porn} or \textit{hentai} are discarded.
    \item We automatically filter out persona image-sentence pairs and dialogues that contain offensive words,
        as introduced in Appendix \ref{subsec:dialog_filtering}.
\end{enumerate}

The above protection schemes can effectively reduce the probability of including \textit{personally identifiable information} (PII) or NSFW in \datasetName, but we cannot guarantee a zero possibility.
Hence, we \textbf{manually checked and excluded} any images containing PII or NSFW content prior to the public release of \datasetName.
Out of 153K images, only 0.6\% (938 images) were filtered out.
To provide further details, 364 images contained face information, 8 images contained NSFW content, and 580 images contained license plate information.
Note that our filtering process was thorough, going as far as excluding images with partially visible faces or reflections caused by glasses in the case of face detection.
Similarly, we eliminated images with unidentifiable plates due to high vehicle speed or low image quality.

\textbf{Consent}.
The consent of Reddit users to collect their data is achieved
through the Reddit API Terms of Use, based on which users expect that
their posts will be publicly available on Reddit and can be
downloaded through Reddit API. However, they do not explicitly agree
on data usage of \datasetName and any related research. To mitigate
this issue, we only distribute URLs instead of images. We also have an
official request form that Reddit users can ask us for data
removal. Furthermore, our data's commercial and for-profit uses 
are restricted -- it should be only available for academic purposes.

\textbf{Human annotation}.
During human annotation, all workers have agreed to the statement of consent
prohibiting personal use of the data shown to them.
Also, they have agreed to comply with the Reddit User Agreement and Privacy Policy
and the Reddit API Terms of Use.

We ensured that our annotators were paid a fair wage of approximately \$16/hour, which is higher than the minimum wage in the countries where we recruited annotators from.
The time to complete each task was determined as 15 seconds by running multiple trials with researchers, and the payment per task was then calculated as \$ 0.07 from this time.
Overall the cost per datapoint was approximately \$0.21.

\section{Further Analyses on \datasetName}
\label{subsec:futher_analyses_mpchat}

\begin{figure}[t]
\begin{center}
    \includegraphics[width=\columnwidth]{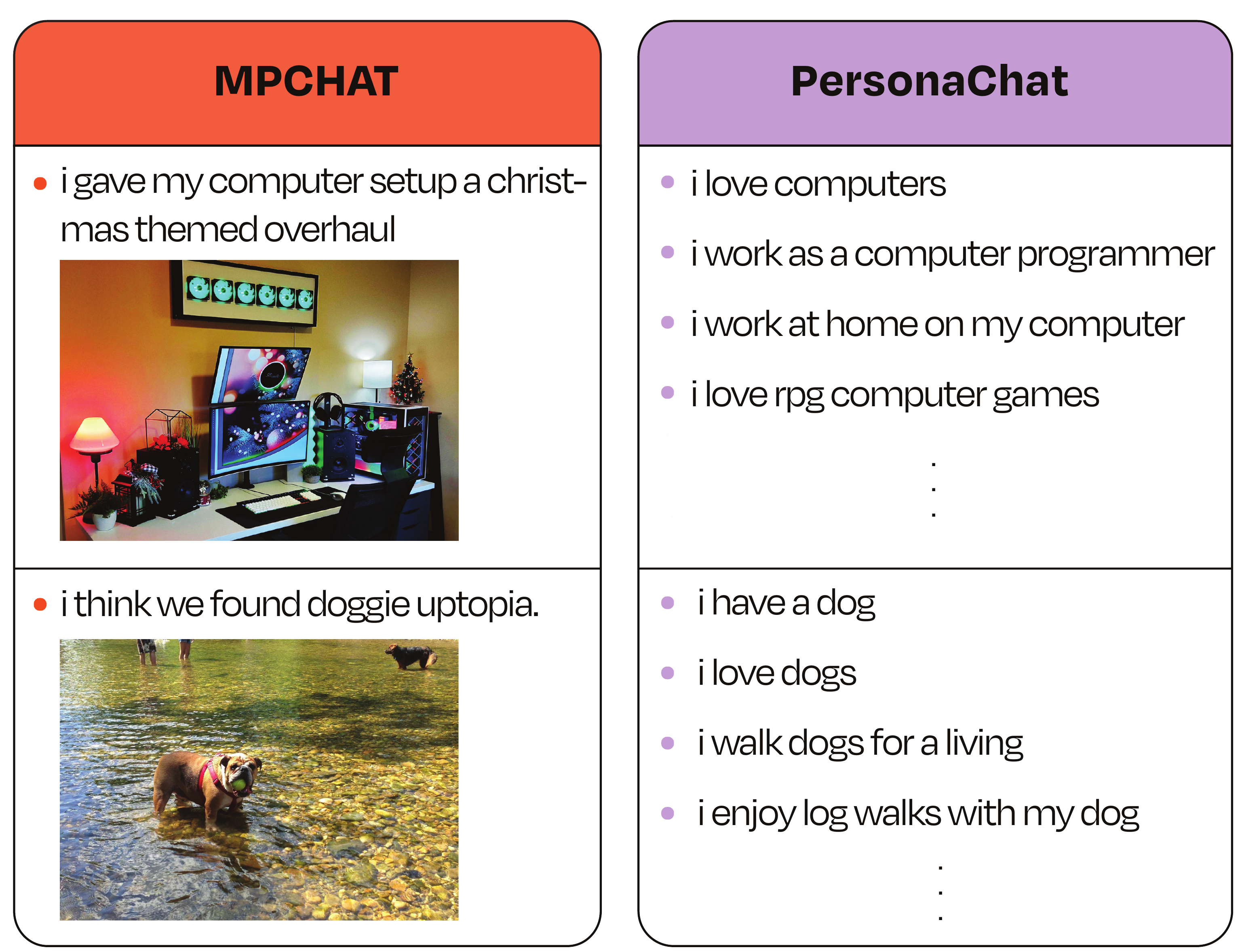}
    \caption{
        Multimodal personas from \datasetName describe episodic memories of personal experiences (\eg computer setup at a Christmas, playing with a dog in water) with visual details,
        while textual personas from PersonaChat reveal personal facts (\eg working as a computer programmer, raising a german shepherd dog).
    }
    \label{fig:persona_comparison}
\end{center}
\end{figure}
\subsection{Comparing Persona in \datasetName and PersonaChat}
Figure~\ref{fig:persona_comparison} shows examples of persona of each dataset: \datasetName and PersonaChat.
Persona in ours reveal one's episodic memory, such as a computer setup at Christmas or playing with a dog in the water.
Furthermore, persona images provide visual information that complements textual information.

\subsection{The Role of Persona Images for \entailmentLabelEntailed Case}
\label{subsec:role_of_persona_image_entailed}
\begin{figure}[t]
\begin{center}
    \includegraphics[width=\columnwidth]{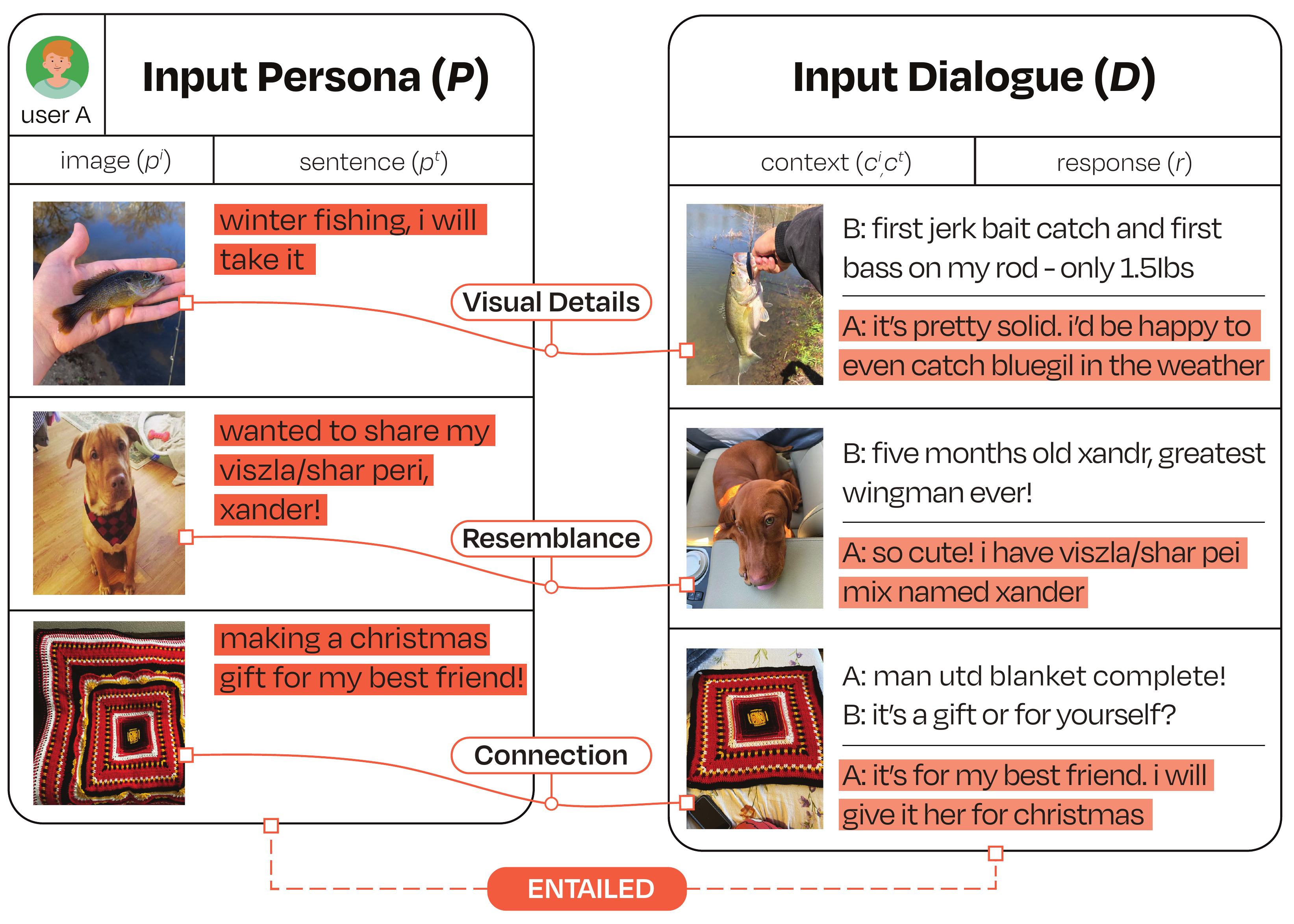}
    \caption{
        Examples of each category, \textit{visual details, resemblance} and \textit{connection}, where visual information of multimodal persona plays an important role in \entailmentLabelEntailed case.}
    \label{fig:example_visual_information}
\end{center}
\end{figure}

In many persona-response pairs, visual information of multimodal persona plays a decisive role in entailing (or not entailing) the response.
In Figure~\ref{fig:example_visual_information}, we categorize the roles of persona images into three: \textit{visual detail}, \textit{resemblance}, and \textit{connection}.

In the first example of Figure~\ref{fig:example_visual_information},
the persona speaker is satisfied with a small-sized fish because it is winter,
leading to the response $r$ that reflects this mindset (referred to as \textit{visual details}).
Regarding the \textit{resemblance} category, the visual information in the persona $p^i$ demonstrates the similarity between the subjects portrayed in the persona and the dialogue, primarily pets or possessions.
In the second example of Figure~\ref{fig:example_visual_information},
the response $r$ exhibits the persona speaker's inquiry about whether the dog in the context image $c^i$ is a vizsla breed, while a vizsla dog is depicted in the persona image $p^i$.
In this case, the speaker's recollection of the dog's appearance becomes the basis for grounding the persona in the utterance.
Lastly, the \textit{connection} category signifies instances when the persona image $p^i$ establishes a connection between the events described in the persona and the dialogue.
It differs from resemblance since the persona speaker and the first speaker in the dialogue are the same,
and thus the same object is presented in both the persona image $p^i$ and the context image $p^i$, such as the identical blanket in the last example of Figure~\ref{fig:example_visual_information}.

\subsection{The Role of Persona Images for \entailmentLabelNotEntailed Case}
\label{subsubsec:categorizing_role_of_persona_images_not_entailed}

\begin{figure}[t]
    \begin{center}
        \includegraphics[width=\linewidth]{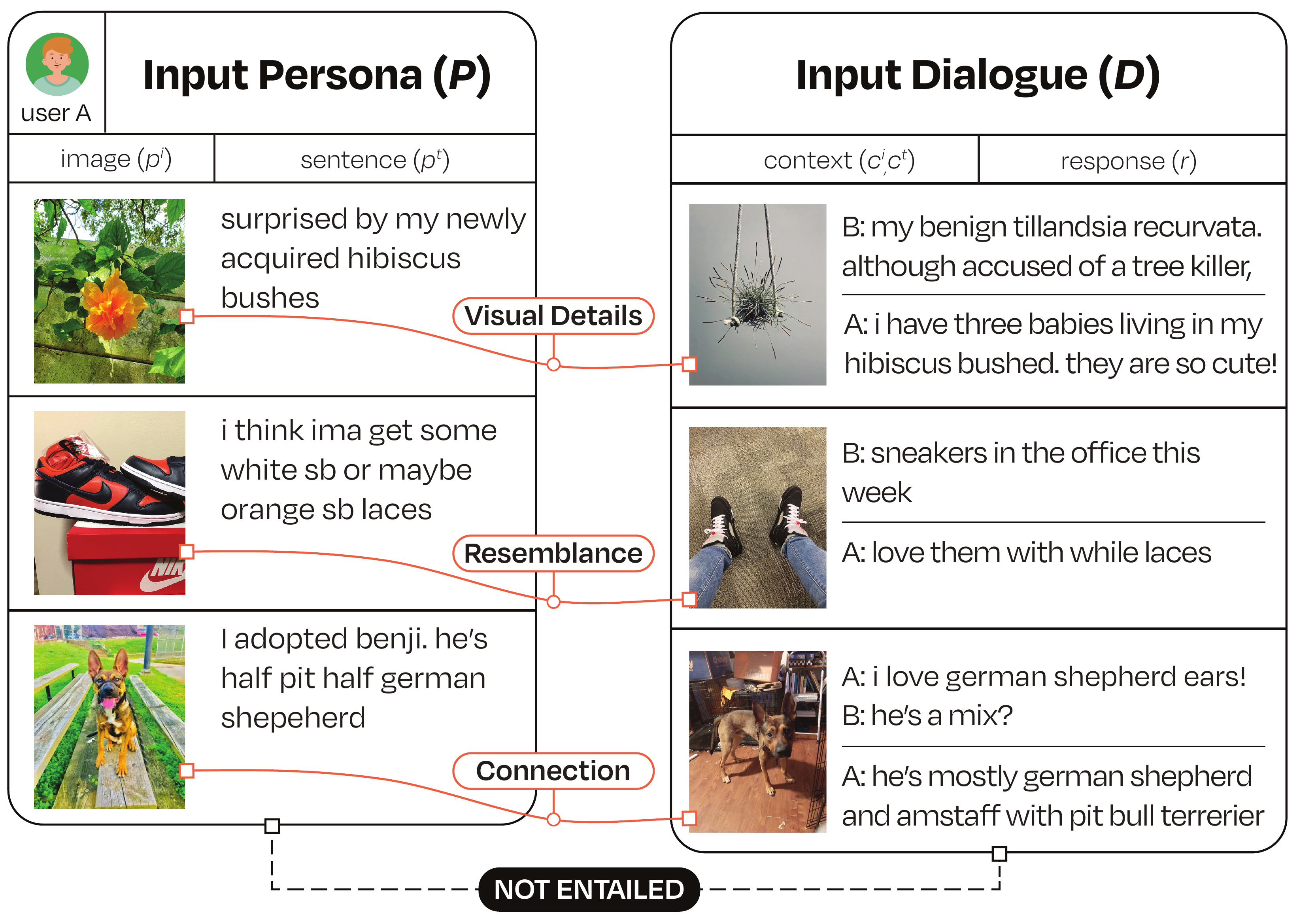}
        \caption{
        Examples of each category, \textit{visual details, resemblance} and \textit{connection}, where visual information of multimodal persona plays a crucial role in \entailmentLabelNotEntailed case.}
        \label{fig:example_visual_information_ne}
    \end{center}
\end{figure}

We classify the role of visuals in multimodal persona into three categories: \textit{visual detail}, \textit{resemblance}, and \textit{connection}.
Figure~\ref{fig:example_visual_information_ne} presents examples for each category, specifically for the \entailmentLabelNotEntailed case.
The criteria for each category are similar to those explained in
Appendix \ref{subsec:role_of_persona_image_entailed}, with slight variations: some persona-response pairs of \textit{connection} category, if \entailmentLabelNotEntailed,
may not have the same object in context and persona images.
The distinction from the \textit{resemblance} category lies solely in whether the persona speaker and the first speaker in the dialogue are the same.
Note that a comprehensive understanding of both images and text is crucial in the \textit{visual detail} category,
as persona images may contain textual information, that is present only in the response but not in the persona sentence, in the form of visuals.
Conversely, the \textit{resemblance} and \textit{connection} categories rely on the presence of resemblance or sameness between the context and persona images as evidence of entailment.

In the first example of Figure~\ref{fig:example_visual_information_ne}, the mention of a specific plant, tillandsia recurvata, is exclusive to the response (\ie three babies) and absent in the persona.
If the plant had appeared in the persona image, which is related to \textit{visual detail}, it would have been labeled as \entailmentLabelEntailed.
The second example, categorized as \textit{resemblance}, is labeled as \entailmentLabelNotEntailed because
the sneakers in the context image do not resemble those in the persona image.
Lastly, the visual information in the final example is critical in labeling it as \entailmentLabelNotEntailed.
The presence of the same dog in both the context and persona images leads to a contradiction,
and since the speaker with the persona is the first speaker in the dialogue, it unquestionably falls into the \textit{connection} category.

{\renewcommand{\arraystretch}{1.0}
    \begin{table}[t!] \begin{center}
    \small
    \begin{tabular}{lccc}
        \toprule
               & \makecell{Train}    & \makecell{Valid}     & \makecell{Test}  \\
        \midrule
        \# dialogue                & 11,975        & 1,516         & 1,509 \\
        \# Speaker                 & 21,197        & 2,828         & 2,797 \\
        \# Utterance               & 34,098        & 4,189         & 4,244 \\
        \midrule
        \# Psn.Speaker             & 8,891         & 1,193          & 1,162 \\
        \# Psn.Response            & 19,048        & 2,303          & 2,321 \\
        \# Gnd.Response            & 6,628         & 709            & 676 \\
        \midrule
        \# Avg.Persona             & 15.89          & 25.6          & 30.76 \\
        \# Avg.Subreddits          & 4.2            & 5.97          & 5.88 \\
        \midrule
        Avg.Utterance.Len       & 18.39             & 18.74          & 19.05 \\
        Avg.Persona.Len         & 10.16             & 10.23          & 10.02 \\
        \bottomrule
    \end{tabular}
    \caption{
        Statistics of our \datasetName in detail.
        \# Psn.Speaker is the number of speakers with multimodal persona.
        \# Psn.Response is the number of responses of persona speakers.
        \# Gnd.Response is the number of responses grounded on the specific persona image-sentence pair.
        \# Avg.Persona is the average number of persona pairs per persona speaker.
        \# Avg.Subreddits indicates the average number of subreddits, from which persona is collected, per persona speaker.
        Avg.Utterance/Persona.Len are the average length of utterances and persona sentences.
    }
    \label{tab:mpchat_stats}
\end{center}\end{table}}

{\renewcommand{\arraystretch}{1.0}
    \begin{table}[t!] \begin{center}
    \begin{adjustbox}{width=\columnwidth}
    \begin{tabular}{lccccc}
        \toprule
        \makecell[l]{Dataset}                                & \makecell{\# Unique\\dialog}   & \makecell{Utterance\\length}  & \makecell{Persona\\type}      & \makecell{Persona\\modality} & \makecell{\#Unique\\image}  \\
        \midrule
        \makecell[l]{PhotoChat}         & 12K                   & 6.3                           & -                             & -                            & 11K                \\
        \makecell[l]{IGC}               & 13K                   & 8.6                           & -                             & -                            & 13K                \\
        \makecell[l]{MMDD}              & 26K                   & 12.0                          & -                             & -                            & 13K                \\
        \makecell[l]{OpenViDial}        & 79K                   & 7.6                           & -                             & -                            & 1.1M               \\
        \makecell[l]{VisualDialog}      & 120K                  & 4.0                           & -                             & -                            & 120K               \\
        \makecell[l]{MMChat}            & 121K                  & 8.5                           & -                             & -                            & 204K               \\
        \makecell[l]{ImageChat}         & 202K                  & 12.3                          & -                             & -                            & 202K               \\
        \midrule
        \makecell[l]{\datasetName}      & 15K                   & 18.5                          & \makecell{Episodic\\memory}   & V,T                          & 153K               \\
        \bottomrule
    \end{tabular}
    \end{adjustbox}
    \caption{
        Comparison of \datasetName with other image-grounded dialogue datasets: PhotoChat~\citep{Zang:2021:ACL}, IGC~\citep{Mostafazadeh:2017:IJCNLP}, MMDD~\citep{Lee:2021:ACL}, MMChat~\citep{Zheng:2021:LREC}, OpenViDial~\citep{Meng:2020:arXiv}, VisualDialog~\citep{Das:2017:CVPR} and ImageChat~\citep{Shuster:2020:ACL_IMAGECHAT}.
        V and T denote visual and textual modality.
    }
    \label{tab:image_dataset_stats}
\end{center}\end{table}}

\subsection{Statistics of \datasetName}
\label{subsubsec:mpchat_stats}
Table~\ref{tab:mpchat_stats} summarizes the statistics of \datasetName.
Thanks to Reddit's abundant sources, the average number of persona image-sentence pairs per user is more than 14.
Table~\ref{tab:image_dataset_stats} compares \datasetName with other image-grounded dialogue datasets.
Only \datasetName deals with multimodal persona consisting of both sentences and images.
Despite the similar number of dialogues, the total number of unique images is larger in \datasetName than in PhotoChat, IGC, MDD and VisualDialog.
Furthermore, the average response length of \datasetName is the largest among other image-grounded dialogue datasets.

\subsection{Persona-groundedness of \datasetName via Subreddit Overlap}
During \datasetName dataset collection, we automatically and manually filtered no persona-related conversations (\S~\ref{subsec:multimodal_entailment}).
To automatically evaluate groundedness of \datasetName between dialogue and multimodal persona, we use \textit{subreddit} metadata as follows.

Every dialogue instance consists of multi-turn dialogue and $m=5$ persona elements of the main speaker.
Then, we check whether any of the subreddits from $m=5$ persona elements is the same as the subreddit from the dialogue.
To sum up, we evaluate persona-groundedness of \datasetName per each filtering process via calculating subreddit overlap ratio (\%):
(1) before automatic filtering, (2) after automatic filtering, and (3) after manual filtering (\ie dialogue instances with \entailmentLabelEntailed labels).
Note that we did not use subreddit metadata during automatic filtering and manual filtering processes (\S~\ref{subsec:multimodal_entailment}),
ensuring that it is fair to use the subreddit overlap ratio as a metric for evaluating persona-groundedness.

For each process, (1) 38.6\%, (2) 68.0\%, and (3) 76.6\% of instances indicate that at least one of the subreddits from persona elements is the same as the subreddit from the dialogue.
The results show that both automatic and manual filtering processes improve subreddit overlap (or persona-groundedness) of \datasetName.

\subsection{Image-Relatedness of \datasetName}
To ensure that dialogues in \datasetName are related to corresponding images ($c^i$),
we conduct a human evaluation on (1) a random subset of \datasetName, (2) randomly paired \datasetName dialogues and images, and (3) a random subset of the ImageChat dataset.
We sample 100 instances per dataset.
For each dataset, 83\%, 19\%, and 72\% of annotators rated that dialogues are related to images,
showing that dialogues in \datasetName are closely related to context images.

\section{Experiment Details}

\subsection{Implementation Details for Three Tasks}
\label{subsec:implementation_details}
In all experiments, we use AdamW optimizer with $\beta_1=0.9, \beta_2=0.999, \epsilon=1e^{-8}$.
We use decoupled weight decay of 0.05 in all experiments.
We do not use linear warmup steps.
We search for the best hyperparameters by testing six different learning rate values ($1e^{-6},2e^{-6},3e^{-6},1e^{-5},2e^{-5},3e^{-5}$).
Regardless of learning rate values, we use a linear scheduler that decreases the learning rate linearly to 0.

We conduct all finetuning experiments on a single NVIDIA Quadro RTX 6000 GPU.
For all experiments, we utilize 13 different random seeds for repeated trials: we then report the average scores and standard deviations.
The number of total parameters for SBERT+ViT, SBERT+CLIP, and CLIP+CLIP models are 376M, 376M, and 366M.

\subsubsection{Next Response Prediction}
We train all models for 5 epochs (approximately 12K steps) with batch size 8.
For SBERT+ViT and SBERT+CLIP, we set learning rate to $1e^{-5}$.
This takes approximately 2.5 GPU hours.
For CLIP+CLIP, we set the learning rate to $3e^{-6}$.
Training this model takes approximately 4 GPU hours.
Note that it takes less time to train SBERT+ViT and SBERT+CLIP than to train CLIP+CLIP
since the image encoder parameters are not updated during training for the former models, whereas they are updated for the latter.

\subsubsection{Grounding Persona Prediction}
In both \texttt{response} and \texttt{no-response} cases, we train all models for 5 epochs (approximately 4K steps) with batch size 8.
For SBERT+ViT and SBERT+CLIP, we set learning rate to $1e^{-5}$.
It takes approximately 1 GPU hour.
For CLIP+CLIP, we set learning rate to $3e^{-6}$, taking approximately 1.5 GPU hours.
Note that the number of total parameters reduces at \texttt{no-response} case:
310M, 310M and 303M for SBERT+ViT, SBERT+CLIP and CLIP+CLIP.

\subsubsection{Speaker Identification}
All models are trained over a period of 5 epochs, which is equivalent to approximately 7.5K steps, using a batch size of 8.
For SBERT+ViT and SBERT+CLIP, we set learning rate to $1e^{-5}$ and $2e^{-5}$ each which takes approximately 4 GPU hour.
As for the CLIP+CLIP, the learning rate is set at $3e^{-6}$, and it takes roughly 5 GPU hours to complete the training.

\subsection{Licenses}
\label{licenses}

We state the licenses that we used, corresponding to the code and models used in this study.
First, we used codes that are distributed under 
\begin{enumerate}
    \item MIT license: 
    CLIP\footnote{\url{https://github.com/openai/CLIP/blob/main/LICENSE}},
    RetinaFace\footnote{\url{https://github.com/biubug6/Pytorch_Retinaface/blob/master/LICENSE.MIT}} \footnote{\url{https://github.com/redcaps-dataset/pytorch-retinaface/blob/master/LICENSE.MIT}}
    InceptionV3\footnote{\url{https://github.com/GantMan/nsfw_model/blob/master/LICENSE.md}}
    \item Apache license 2.0: 
    ViT, BERT \footnote{\url{https://github.com/huggingface/transformers/blob/v4.17.0/LICENSE}}
    \item BSD 3-Clause "New" or "Revised" License:
    ALBEF\footnote{\url{https://github.com/salesforce/ALBEF/blob/main/LICENSE.txt}},
    BLIP\footnote{\url{https://github.com/salesforce/BLIP/blob/main/LICENSE.txt}},
    X-VLM\footnote{\url{https://github.com/zengyan-97/X-VLM/blob/master/LICENSE}}
\end{enumerate}

We could not find the license for the license plate detection code, but the code was from a public GitHub repository.
Also, Yolo v3, used in license plate detection, has a GNU General Public License v3.0
\footnote{\url{https://github.com/ultralytics/yolov3/blob/master/LICENSE}}.
Since all the licenses include permissions for commercial use, modification, distribution, patent use, and private use of the artifacts, 
we comply with the regulations of the above licenses.

\section{Further Analyses on Experiments}
\label{sec:further_analyses_experiments}

\subsection{Ablation Study based on Textual Persona-Response Similarity}
Previously, we observed that conditioning on persona sentences yielded better performance compared to conditioning on persona images
in the next response prediction (\S~\ref{subsec:experiment_mprs}) and the speaker identification (\S~\ref{subsec:experiment_si}) tasks.
We hypothesize that dialogue models tend to retrieve responses based on textual similarities, such as lexical or semantic similarity, between the response $r$ and persona sentences $P^t$.
Conversely, we assume that dialogue models face challenges in retrieving responses (or speakers) when this textual similarity is low, where persona images $P^i$ may contain useful hints.

To investigate the importance of persona images in specific dialogue instances, we split the test set as follows:
for each instance, we calculate F1 score between the response $r$ and persona sentences $P^t=\{p^t_1,...,p^t_m\}$: F1$^r_{t_1}$,...,F1$^r_{t_m}$.
We then identify the maximum F1 value and split them using a specific threshold (\ie 0.3).
We refer to dialogue instances with lower F1 scores as the \texttt{low-f1} subset, while the remaining instances form the \texttt{high-f1} subset.
In the next response prediction task (or the speaker identification task), the \texttt{low-f1} subset contains 571 (or 284) instances, while the \texttt{high-f1} subset consists of 1,750 (or 1,255) instances.
For each subset, we measure the performance gap between dialogue models with full inputs and models without persona images, as shown in Table~\ref{tab:ablation}.

{\renewcommand{\arraystretch}{1.0}
    \begin{table}[t!]
    \begin{center}
    \begin{adjustbox}{width=\columnwidth}
    \begin{tabular}{llll}
        \toprule
                                         & \makecell{SBERT+ViT}    & \makecell{SBERT+CLIP}  & \makecell{CLIP+CLIP}  \\
        \midrule
        \multicolumn{4}{l}{\textbf{Next Response Prediction (\texttt{high-f1})}}     \\
        $c,P^t$                                 & 67.89             & 68.29  & 74.25          \\
        $c,P$ (Full)                            & 69.39             & 68.86   & 74.55          \\
        $\Delta$                                & +1.5 & +0.57 & +0.3 \\
        \multicolumn{4}{l}{\textbf{Next Response Prediction (\texttt{low-f1})}}     \\
        $c,P^t$                                 & 52.25             & 51.49     & 65.62          \\
        $c,P$ (Full)                            & 54.53             & 54.64     & 67.66          \\
        $\Delta$                                & \textbf{+2.28}    & \textbf{+3.15}     & \textbf{+2.04} \\
        \midrule
        \multicolumn{4}{l}{\textbf{Speaker Identification (\texttt{high-f1})}}     \\
        $c,r,\mathbb{P}_{c}^{t}$                & 59.7             & 59.15   & 61.69          \\
        $c,r,\mathbb{P}_{c}$ (Full)             & 58.86             & 59.59   & 62.77          \\
        $\Delta$                                & -0.84 & +0.44 & +1.08 \\
        \multicolumn{4}{l}{\textbf{Speaker Identification (\texttt{low-f1})}}     \\
        $c,r,\mathbb{P}_{c}^{t}$                & 45.19             & 46.71   & 53.76          \\
        $c,r,\mathbb{P}_{c}$ (Full)             & 49.53             & 49.76   & 58.69          \\
        $\Delta$                                & \textbf{+4.34} & \textbf{+3.05} & \textbf{+4.93} \\
        \bottomrule
    \end{tabular}
    \end{adjustbox}
    \caption{Ablation study focused on textual persona-response similarity in two tasks: the next response prediction and the speaker identification.
        For each subset (referred to as \texttt{high-f1} and \texttt{low-f1}) within each task, we measure the performance gap (denoted as $\Delta$ of R@1) between the models with full inputs and the models without persona images.
        In both tasks, we observe larger performance gaps $\Delta$ in the \texttt{low-f1} subsets.
    }
    \label{tab:ablation}
\end{center}\end{table}}

\textbf{All models perform better in the \texttt{high-f1} subsets compared to the \texttt{low-f1} subsets.}
In both tasks, the models demonstrate improved performance in the \texttt{high-f1} subsets compared to the \texttt{low-f1} subsets,
providing evidence that persona sentences $P^t$ are utilized as valuable cues for predicting the response or speaker.

\textbf{The performance gaps are more pronounced in the \texttt{low-f1} subsets than in the \texttt{high-f1} subsets.}
The performance gaps between the models with full inputs and the models without persona images are larger in the \texttt{low-f1} subsets.
This indicates that textual information from persona sentences tends to be less helpful,
while visual information from persona images $P^i$ becomes crucial for predicting the gold response or speaker in such cases.

In conclusion, persona images play a critical role, particularly when persona sentences fail to provide useful cues for predicting the responses or speakers.

\subsection{Ablation Study on Persona-Consistency in Grounding Persona Prediction Task}
Grounding persona prediction task is designed to ensure both multimodal context-awareness and multimodal persona-consistency, as mentioned in \S~\ref{sec:task_definition}.
We focus on evaluating multimodal persona-consistency by excluding context information as shown in Table~\ref{tab:gpp_persona_consistency}.

{\renewcommand{\arraystretch}{1.0}
    \begin{table}[t!]
    \begin{center}
    \small
    \begin{tabular}{lll}
        \toprule
        Model                                 & \makecell{R@1$\uparrow$}    & \makecell{MRR$\uparrow$}    \\
        \midrule
        \multicolumn{3}{l}{\textbf{CLIP+CLIP}}     \\
        $\bar{P}^i$     & 53.82$\pm$1.11             & 63.72$\pm$0.82             \\
        $\bar{P}^t$     & 43.82$\pm$1.33             & 54.57$\pm$0.87             \\
        $\bar{P}$ & \textbf{56.18$\pm$1.44}$^{**}$    & \textbf{66.11$\pm$0.97}$^{**}$    \\
        \midrule
        \textcolor{gray}{$c,\bar{P}$ (Full)} & \textcolor{gray}{82.32$\pm$0.75}    & \textcolor{gray}{88.52$\pm$0.46}    \\
        \textcolor{gray}{$c,r,\bar{P}$ (Full)}  & \textcolor{gray}{94.79$\pm$0.5}    & \textcolor{gray}{96.94$\pm$0.28}    \\
        \bottomrule
    \end{tabular}
    \caption{Results of the grounding persona prediction task on the CLIP+CLIP model without context and without response information.
      Symbols means $c$: context text and image, $\bar{P}^i$: remainder persona images, $\bar{P}^t$: remainder persona sentences, and $\bar{P}=\bar{P}^i \cup \bar{P}^t$.
      We report the average scores with standard deviations.
      Asterisks denote statistical significance of differences between \textit{full} model and its closest competitor  (*p < 0.05 and **p < 0.001).
      Note that models with context information are highlighted in \textcolor{gray}{gray} and serve as upper-bound models in \texttt{response} or \texttt{no-response} cases.
    }
    \label{tab:gpp_persona_consistency}
\end{center}\end{table}}

\textbf{Omitting context information significantly lowers performance.}
Models without $c$ perform worse compared to models with either $c,\bar{P}$ or $c,r,\bar{P}$, highlighted in gray.
This result highlights the crucial role of context information in the grounding persona prediction task.
Nevertheless, models without $c$ can still achieve a recall rate of over 50\% in predicting the persona element $p^*$ at Recall@1, showing the task's persona-consistent characteristics.

\textbf{Still, using both remainder persona images $\bar{P}^i$ and persona sentences $\bar{P}^t$ maximizes performance.}
Models equipped with both $\bar{P}^i$ and $\bar{P}^t$ achieve the highest scores in terms of Recall@1 and MRR scores,
indicating the importance of leveraging multimodal persona information to its full extent.
In addition, note that the results indicate that $\bar{P}^i$ contributes more signifcantly to model improvement compared to $\bar{P}^t$.

In summary, the results illustrate the grounding persona prediction task's ability to capture persona-consistent traits.
That is, the model exhibits the capability to predict persona element $p^*$ by only leveraging the remainder persona set $\bar{P}$.

\subsection{Multimodal Persona Entailment Task}
Beyond retrieval-based dialogue tasks (\ie next response prediction, grounding persona prediction, and speaker identification),
we propose an additional task to predict persona-response entailment, whose labels were collected via human annotations in~\S~\ref{subsec:multimodal_entailment}.

The goal of the task is to predict the two-class entailment label $\Pr(e|c,r,p)$, given a context $c=(c^i,c^t)$, a response $r$, and a persona image-sentence pair $p=(p^i,p^t)$.

\subsubsection{Models}
We use ALBEF~\citep{Li:2021:NeurIPS}, BLIP~\citep{Li:2022:ICML} and X-VLM~\citep{Zeng:2022:ICML} for the task, since they are the state-of-the-art multimodal representation models.
We have chosen to use the variant of each model that was utilized for the NLVR2~\citep{Suhr:2019:ACL} task as is,
in order for the models to be able to handle both two images (\ie $c^i,p^i$) and two text components (\ie $c^t,p^t$) in a single example.
To be specific, we concatenate $c^t,p^t$ in a single sequence and provide it with $c^i,p^i$ to each model as inputs.

For ALBEF, we set the learning rate to 3e-05.
For BLIP, we set the learning rate to 1e-05.
For X-VLM, we set the learning rate to 2e-05.
We use AdamW optimizer with $\beta_1=0.9, \beta_2=0.999, \epsilon=1e^{-8}$.
We use decoupled weight decay of 0.05 in all experiments.
We do not use linear warmup steps.
We use a linear scheduler that decreases the learning rate linearly to 0.
For all models, we train them for 5 epochs with batch size 8.
This takes approximately 2 GPU hours on a single NVIDIA Quadro RTX 6000 GPU for each model.

For the evaluation metric, we measure the accuracy score.

\subsubsection{Experiments}
We compare models that use full inputs with baselines that only make use of partial inputs.

\textbf{Baselines.}
We utilize the following baselines.
(1) \textit{Dialogue} ($c,r$) + \textit{speaker's persona sentence} ($p^t$):
Instead of providing the speaker's persona image $p_i$, each model is given a random image with an attention mask of value 0.
(2) \textit{Dialogue} ($c,r$) + \textit{speaker's persona image} ($p^i$):
We provide only response $r$ to models without the speaker's persona sentence $p^t$.

\textbf{Results.}
According to Table~\ref{tab:me}, all models with persona images show better performances than those with persona sentences.
It means that persona images, as well as persona sentences, help the model to improve the persona entailment task.
Combining persona images and sentences achieves the best performance for all models,
showing that the task requires multimodal reasoning ability as intended.

{\renewcommand{\arraystretch}{1.0}
    \begin{table}[t!]
      \begin{center}
       \small
    \begin{tabular}{lll}
        \toprule
        Model                                       & \makecell{Test set Acc.$\uparrow$}    \\
        \midrule
        \multicolumn{2}{l}{\textbf{BLIP}}     \\
        $c,p^t$         & 69.02$\pm$3.05              \\
        $c,p^i$          & 70.46$\pm$0.62                \\
        $c,p$ (Full)      & \textbf{72.42$\pm$0.64}$^{**}$       \\
        \midrule
        \multicolumn{2}{l}{\textbf{X-VLM}}     \\
        $c,p^t$         & 68.52$\pm$4.03              \\
        $c,p^i$          & 71.48$\pm$1.43                \\
        $c,p$ (Full)      & \textbf{74.13$\pm$2.63}$^{*}$       \\
        \midrule
        \multicolumn{2}{l}{\textbf{ALBEF}}     \\
        $c,p^t$         & 67.47$\pm$4.55              \\
        $c,p^i$          & 70.67$\pm$1.71                \\
        $c,p$ (Full)      & \textbf{72.47$\pm$4.33}       \\
        \bottomrule
    \end{tabular}
    \caption{Results of the multimodal persona entailment task. Symbols means $c^t$: context text, $c^i$: context image, $p^i$: persona image, and $p^t$: persona sentence. Also, $c= c^t \cup c^i $ and $p = p^i \cup p^t$. 
        We report the average scores with standard deviations.
        Asterisks denote statistical significance of differences between \textit{full} model and its closest competitor  (*p < 0.05 and **p < 0.001).
    }
    \label{tab:me}
\end{center}\end{table}}

\section{Coverage of Domains}
For both the text and image data in \datasetName, their coverage of domains is a subset of Reddit posts.
To be more precise, the content of \datasetName is derived from subreddits listed in Appendix~\ref{subsec:persona_subreddit_list} and Appendix~\ref{subsec:dialog_subreddit_list}.

\subsection{List of all subreddits for personas}
\label{subsec:persona_subreddit_list}
We list all subreddits curated for multimodal persona collection.
There are 648 subreddits for all multimodal personas, consisting of 140,658 image-sentence pairs,
including 16,327 pairs used to obtain persona entailment labels.

{\small
pics (7274), cats (7172), aww (6785), succulents (5372), houseplants (4957), gardening (4805), crochet (4135), baking (3275),
aquariums (3018), food (2489), sneakers (2069), somethingimade (2018), foodporn (1885), mildlyinteresting (1576), breadit (1489), thriftstorehauls (1431), rabbits (1398), fountainpens (1341), crafts (1293), guineapigs (1293), bicycling (1204), woodworking (1171), embroidery (1142), blackcats (1135), quilting (1118), cakedecorating (1107), dogpictures (1097), bladesmith (1094), plantedtank (1016), bettafish (984), knives (946), indoorgarden (875), knitting (828), crossstitch (819), coins (810), blacksmith (806), trees (748), plantclinic (744), cactus (737), squirrels (714), catpictures (680), rarepuppers (669), itookapicture (658), parrots (642), redditlaqueristas (621), mechanicalkeyboards (604), earthporn (602), orchids (597), sewing (590), plants (577), castiron (570), corgi (569), tea (565), proplifting (551), pitbulls (550), tonightsdinner (550), snakes (549), fishing (543), sourdough (533), photocritique (533), husky (515), eyebleach (498), beerporn (487), horses (475), hotpeppers (470), spiders (465), reptiles (453), mycology (445), knifeclub (439), shittyfoodporn (419), beardeddragons (405), knifemaking (394), brochet (391), germanshepherds (368), pizza (355), watches (353), silverbugs (345), shrimptank (343), flyfishing (340), lookatmydog (328), backyardchickens (327), bulldogs (324), casualknitting (318), pottery (311), crystals (303), cakewin (298), cocktails (298), birding (292), smoking (274), vinyl (266), vegetablegardening (262), dachshund (258), hamsters (255), guns (246), hiking (245), flowers (243), campingandhiking (241), cookiedecorating (241), bbq (238), savagegarden (237), equestrian (236), vegan (232), chickens (226), bonsai (221), grilling (220), birdpics (219), airplants (218), supermodelcats (217), lego (213), diy (209), tools (206), barista (205), tarantulas (205), reeftank (205), eatsandwiches (204), ceramics (199), trucks (196), camping (193), duck (192), amigurumi (191), yarnaddicts (191), drunk (188), pyrex\_love (185), spaceporn (183), bulletjournal (182), spiderbro (180), carporn (178), spicy (177), subaru (176), cozyplaces (176), 3dprinting (175), wirewrapping (175), fixedgearbicycle (174), dessertporn (172), battlestations (170), bikecommuting (169), chihuahua (167), edc (165), steak (163), cheesemaking (161), catloaf (160), natureisfuckinglit (156), pugs (156), metaldetecting (156), floof (155), interestingasfuck (154), gamecollecting (154), homestead (152), rats (151), zerowaste (151), haworthia (150), tuxedocats (149), mineralporn (149), kayaking (147), rainboweverything (144), burgers (142), 1200isplenty (135), pomeranians (135), miata (134), monstera (134), outdoors (134), modelmakers (134), insects (131), leathercraft (129), tuckedinkitties (128), travel (128), flytying (128), jeep (127), goldenretrievers (125), sailing (125), herpetology (124), cat (121), curledfeetsies (121), cakes (121), bassfishing (121), journaling (120), chefknives (118), frogs (118), greatpyrenees (117), metalworking (115), delightfullychubby (115), turning (114), macarons (113), leopardgeckos (113), microgrowery (112), marijuanaenthusiasts (111), kitting (110), penmanshipporn (110), christmas (109), sneks (108), mid\_century (108), plantidentification (108), vans (107), autos (105), sonyalpha (103), handwriting (102), rockhounds (102), pens (100), fermentation (100), mealprepsunday (97), exposureporn (96), ferrets (95), hunting (95), veganfoodporn (95), terrariums (95), plantsandpots (95), hoyas (93), golf (91), astrophotography (91), torties (90), justrolledintotheshop (90), beginnerwoodworking (90), watchescirclejerk (89), vintageaudio (89), mostbeautiful (88), takeaplantleaveaplant (88), doggos (88), upcycling (86), catbellies (86), entomology (85), wildlifephotography (84), bostonterrier (83), ramen (83), astronomy (83), funkopop (82), cockatiel (82), sushi (81), wicked\_edge (81), woodcarving (81), 4runner (81), ballpython (80), randomactsofpolish (80), longboarding (79), antiques (77), muglife (76), botanicalporn (76), chonkers (76), seniorkitties (75), awww (75), aviation (75), gunpla (75), jigsawpuzzles (74), crestedgecko (73), lithops (73), awwnverts (73), hotsauce (72), goldfish (72), bmw (72), needlefelting (71), foraging (71), jewelrymaking (71), canning (70), veganrecipes (70), classiccars (70), 4x4 (69), homebrewing (69), vegetarian (69), damnthatsinteresting (69), jewelry (68), aquaticsnails (68), sousvide (68), amateurphotography (68), bordercollie (68), weed (67), amateurroomporn (67), welding (67), dessert (67), crh (66), seriouseats (65), vandwellers (65), whiskey (63), siberianhusky (63), mustang (63), beagle (63), kayakfishing (62), plant\_progress (62), mead (62), covidcookery (61), drunkencookery (61), budgies (61), skyporn (60), puppysmiles (59), snails (59), catsareassholes (59), chinesefood (59), beforenafteradoption (59), fishing\_gear (59), australiancattledog (59), cottagecore (59), panporn (58), roses (58), shiba (58), projectcar (58), workbenches (58), labrador (57), turtle (57), oldmandog (56), dumpsterdiving (56), charcuterie (55), analog (55), airsoft (55), siamesecats (55), audiophile (54), ar15 (53), knifeporn (53), swords (53), ntbdbiwdfta (53), jarrariums (53), geckos (53), illegallysmolcats (52), bakingnoobs (52), cupcakes (52), nails (52), vintage (52), australianshepherd (52), skiing (52), breakfastfood (51), hotwheels (51), mushrooms (51), climbing (51), birdsofprey (51), landscaping (51), pourpainting (51), pothos (51), hedgehog (50), grilledcheese (50), cichlid (50), polymerclay (50), cheese (50), healthyfood (50), dunksnotdead (50), kitchenconfidential (49), abandonedporn (49), beekeeping (49), wildernessbackpacking (49), discgolf (49), aquascape (49), superbowl (48), honda (47), propagation (47), shrooms (47), origami (46), aquarium (46), multicopter (46), malelivingspace (45), ford (45), macroporn (45), dvdcollection (45), butterflies (44), xbiking (44), functionalprint (44), flashlight (44), cityporn (43), volkswagen (43), bikesgonewild (43), gshock (43), bushcraft (42), cricut (42), matureplants (42), lockpicking (42), ketorecipes (42), gardenwild (42), bees (41), animalporn (41), retrogaming (41), interiordesign (40), stance (40), harley (40), aldi (40), volvo (40), guitarpedals (40), drums (39), toyotatacoma (39), handtools (39), wine (38), absoluteunits (38), cherokeexj (38), beadsprites (38), slowcooking (38), resincasting (38), vexillology (38), dog (37), drunkknitting (37), foxes (37), pug (37), chameleons (37), visiblemending (36), beerandpizza (36), wigglebutts (36), mini (36), mountainbiking (36), headphones (35), whiskyporn (35), bathandbodyworks (35), espresso (34), pelletgrills (34), soapmaking (34), velvethippos (34), salsasnobs (34), moths (34), axolotls (34), wellworn (33), backpacking (33), cassetteculture (33), waltdisneyworld (33), sanpedrocactus (33), mainecoons (32), whiskeytribe (32), geology (31), blop (31), shihtzu (31), shittyveganfoodporn (31), sharks (31), antkeeping (31), cute (31), homedecorating (31), begonias (31), owls (31), wrangler (31), rolex (31), dobermanpinscher (30), mushroomgrowers (30), greatdanes (30), actionfigures (30), paintball (29), chinchilla (29), catsandplants (29), bookshelf (28), perfectfit (28), roastmycar (28), glocks (28), golfgti (28), porsche (28), retrobattlestations (28), planetzoo (28), canadaguns (28), catswithjobs (27), mazda3 (27), mazda (27), keto\_food (27), kombucha (27), disneyland (27), rccars (27), transformers (27), guitars (27), greyhounds (26), weaving (25), craftbeer (25), buyitforlife (25), budgetaudiophile (25), electricians (25), osha (25), snowboarding (25), catsmirin (25), catsinsinks (25), scotch (24), hometheater (24), composting (24), gunporn (24), glassheads (24), ants (24), teaporn (24), breakfast (23), fish (23), pokemontcg (23), toyota (23), dualsport (23), tastyfood (22), nikon (22), bonecollecting (22), gravelcycling (22), trains (22), bento (22), boxer (22), audi (22), waterporn (21), boating (21), formula1 (21), nebelung (21), bookhaul (20), modeltrains (20), femalelivingspace (20), techsupportgore (19), powerwashingporn (19), soup (19), guitarporn (19), reloading (19), natureporn (19), poodles (19), philodendron (19), typewriters (18), tinyanimalsonfingers (18), archery (18), mechanicalpencils (18), firearms (18), gamingpc (18), carpentry (18), otters (18), scooters (18), vintageapple (18), fordranger (17), tacos (17), cameras (17), subaruforester (17), bernesemountaindogs (17), amiibo (17), cartalk (17), toolporn (17), glutenfree (17), tortoise (17), trailrunning (17), tequila (16), chefit (16), analogcommunity (16), luthier (16), bmx (16), tacobell (16), mantids (16), vhs (16), roomporn (15), fiddleleaffig (15), gameboy (15), macrame (14), designmyroom (14), lizards (14), bookporn (14), bengalcats (14), frenchbulldogs (14), sloths (14), comicbookcollecting (14), hockeyjerseys (14), starwarscollecting (14), instantpot (14), seiko (14), polaroid (14), machinists (14), shroomid (14), coffeestations (13), geologyporn (13), icecreamery (13), wrx (13), hvac (13), ender3 (13), carnivorousplants (13), architectureporn (13), camaro (13), masseffect (13), balisong (13), tamagotchi (13), ft86 (13), farming (12), urbanexploration (12), f150 (12), shroomers (12), permaculture (12), cabinporn (12), beerwithaview (12), ruralporn (12), wewantplates (12), samoyeds (12), sigsauer (12), jdm (12), cornsnakes (12), gold (11), photographs (11), crows (11), nerf (11), rottweiler (11), blender (11), sffpc (11), supremeclothing (11), gemstones (10), homelab (10), pebble (10), longrange (10), villageporn (10), ak47 (10), playingcards (10), tfablineporn (10), mushroomporn (9), jellyfish (9), tiedye (9), winterporn (9), corvette (9), volumeeating (9), liberalgunowners (9), warhammer (8), goldendoodles (8), skateboarding (8), animefigures (8), czfirearms (8), dirtbikes (8), simracing (8), siberiancats (8), averagebattlestations (8), cubers (8), bassguitar (8), budgetfood (7), fireporn (7), streetphotography (7), birdphotography (7), legostarwars (7), vinyljerk (7), regularcarreviews (7), petmice (7), homegym (7), synthesizers (7), motorcycleporn (7), telescopes (6), cider (6), schnauzers (6), fossilporn (6), birds (6), plantbaseddiet (5), tractors (5), awwducational (5), infrastructureporn (5), melts (5), helicopters (5), lightsabers (5), mousereview (5), mercedes\_benz (5), motorcycle (5), unclebens (5), liminalspace (5), seaporn (4), berries (4), houseporn (4), microgreens (4), crtgaming (4), focusst (4), machineporn (4), thedepthsbelow (3), pkmntcgcollections (3), boatporn (3), autumnporn (3), f1porn (3), desksetup (3), microporn (2), nfa (2), squishmallow (2), onewheel (2), bridgeporn (1), desertporn (1), underwaterphotography (1), castles (1), weatherporn (1), workspaces (1)
}

\subsection{List of all subreddits for dialogues}
\label{subsec:dialog_subreddit_list}
We list all subreddits curated for dialogue collection. There are 110 subreddits in total for the 15,000 dialogues.

{\small
pics (1287), cats (1075), cakedecorating (771), bladesmith (472), houseplants (440), gardening (414), itookapicture (400), breadit (363), tonightsdinner (313), crochet (312), succulents (309), bicycling (275), guineapigs (256), aquariums (246), diy (244), mildlyinteresting (226), sneakers (212), rabbits (210), baking (198), crossstitch (186), burgers (182), casualknitting (181), earthporn (180), fountainpens (178), embroidery (172), grilling (171), rarepuppers (167), camping (166), ceramics (163), cocktails (163), blackcats (162), bassfishing (158), tea (152), dogpictures (148), husky (148), cakewin (144), hiking (132), zerowaste (130), cookiedecorating (128), food (125), brochet (118), parrots (113), cheesemaking (109), upcycling (109), plantedtank (109), bikecommuting (107), thriftstorehauls (104), flyfishing (100), corgi (98), crystals (93), snakes (91), mechanicalkeyboards (89), coins (85), horses (77), pitbulls (77), eyebleach (77), chickens (76), squirrels (75), dachshund (73), duck (69), beardeddragons (69), quilting (68), bulldogs (65), germanshepherds (61), foodporn (58), barista (57), pomeranians (55), catpictures (55), reptiles (53), castiron (53), blacksmith (51), kayaking (51), watches (51), indoorgarden (50), greatpyrenees (49), campingandhiking (47), workbenches (47), lookatmydog (43), chinesefood (42), equestrian (40), battlestations (40), sewing (40), photocritique (40), hotpeppers (40), pizza (39), sourdough (37), sailing (36), orchids (36), trucks (35), vinyl (34), plants (33), cozyplaces (33), bettafish (32), cactus (32), beerandpizza (29), spiders (29), charcuterie (24), pug (21), veganrecipes (19), knives (18), doggos (18), amateurphotography (17), mycology (17), fishing (17), villageporn (5), infrastructureporn (2), desertporn (1), awwducational (1), seaporn (1), f1porn (1)
}

\end{document}

%% file: math_commands.tex
\usepackage{amsmath,amsfonts,bm}

\def\eqref#1{equation~\ref{#1}}

\def\1{\bm{1}}

\DeclareMathAlphabet{\mathsfit}{\encodingdefault}{\sfdefault}{m}{sl}
\SetMathAlphabet{\mathsfit}{bold}{\encodingdefault}{\sfdefault}{bx}{n}



%% file: mymacros.tex
\usepackage{color}
\usepackage{amsmath}
\usepackage{amssymb}
\usepackage{multirow, varwidth}
\usepackage{dblfloatfix}
\usepackage{verbatim}
\makeatletter
\newif\if@restonecol
\makeatother

\usepackage[ruled,vlined]{algorithm2e}
\usepackage{xr}
\usepackage[amsmath,thmmarks]{ntheorem}

\DeclareRobustCommand\onedot{\futurelet\@let@token\@onedot}
\def\onedot{. }
\def \ie{i.e.,\xspace}
\def \eg{e.g.,\xspace}